\documentclass[]{arxiv_article}

\usepackage[toc,page,header]{appendix}
\microtypesetup{expansion=false}
\usepackage{xspace}
\usepackage{enumitem}
\usepackage[linesnumbered,ruled,vlined]{algorithm2e}
\usepackage{amsmath}
\usepackage{amssymb}
\definecolor{darkgreen}{rgb}{0.0,0.45,0.0}
\newcommand{\ours}{\textbf{\texttt{ECHO}}\xspace}
\newcommand{\ar}{\textbf{\texttt{ECHO$_{\text{AR}}$}}\xspace}
\newcommand{\base}{\textbf{\texttt{ECHO$_{\text{Base}}$}}\xspace}
\newcommand{\oursblkeight}{\textbf{\texttt{ECHO}}$_{\text{blk8}}$\xspace}
\newcommand{\oursblkfour}{\textbf{\texttt{ECHO}}$_{\text{blk4}}$\xspace}
\newcommand{\baseblkfour}{\textbf{\texttt{ECHO}}$_{\text{Base,blk4}}$\xspace}
\newcommand{\baseblkeight}{\textbf{\texttt{ECHO}}$_{\text{Base,blk8}}$\xspace}

\newcommand{\keywords}[1]{\par\medskip\noindent\textbf{Keywords:}\space\def\and{, }#1}

\title{ECHO: Efficient Chest X-ray Report Generation with One-step Block Diffusion}

\author[1,2,*,\protect\star]{Lifeng Chen}{}
\author[1,3,*,\protect\star]{Tianqi You}{}
\author[1,\dagger,\ddagger]{Hao Liu}{}
\author[1]{Zhimin Bao}{}
\author[1]{Jile Jiao}{}
\author[1]{Xiao Han}{}
\author[1]{Zhicai Ou}{}
\author[1]{Tao Sun}{}
\author[1]{Xiaofeng Mou}{}
\author[2]{Xiaojie Jin}{}
\author[1,\dagger]{Yi Xu}{}

\affiliation[1]{AIRC, Midea Group}
\affiliation[2]{Beijing Jiaotong University}
\affiliation[3]{Dalian University of Technology}

\contribution[*]{Equal Contribution}
\contribution[\dagger]{Corresponding Author}
\contribution[\ddagger]{Project Leader,}
\contributionpar[\star]{Work completed during an internship at Midea AI Research Center.}

\abstract{
Chest X-ray report generation (CXR-RG) has the potential to substantially alleviate radiologists' workload. However, conventional autoregressive vision--language models (VLMs) suffer from high inference latency due to sequential token decoding. Diffusion-based models offer a promising alternative through parallel generation, but they still require multiple denoising iterations. Compressing multi-step denoising to a single step could further reduce latency, but often degrades textual coherence due to the mean-field bias introduced by token-factorized denoisers.
To address this challenge, we propose \textbf{ECHO}, an efficient diffusion-based VLM (dVLM) for chest X-ray report generation. ECHO enables stable one-step-per-block inference via a novel Direct Conditional Distillation (DCD) framework, which mitigates the mean-field limitation by constructing unfactorized supervision from on-policy diffusion trajectories to encode joint token dependencies. In addition, we introduce a Response-Asymmetric Diffusion (RAD) training strategy that further improves training efficiency while maintaining model effectiveness.
Extensive experiments demonstrate that ECHO surpasses state-of-the-art autoregressive methods, improving RaTE and SemScore by \textbf{64.33\%} and \textbf{60.58\%} respectively, while achieving up to \textbf{$8\times$} inference speedup with negligible degradation in clinical accuracy.

}

\date{\today}
\correspondence{Hao Liu, Yi Xu}
\checkdata[Project Page]{\url{https://echo-midea-airc.github.io/}}

\begin{document}
\begingroup
\hypersetup{linkcolor=black}%
\maketitle
\endgroup

\section{Introduction}
\label{sec:intro}

\begin{figure*}[t]
    \centering
    \includegraphics[width=\textwidth]{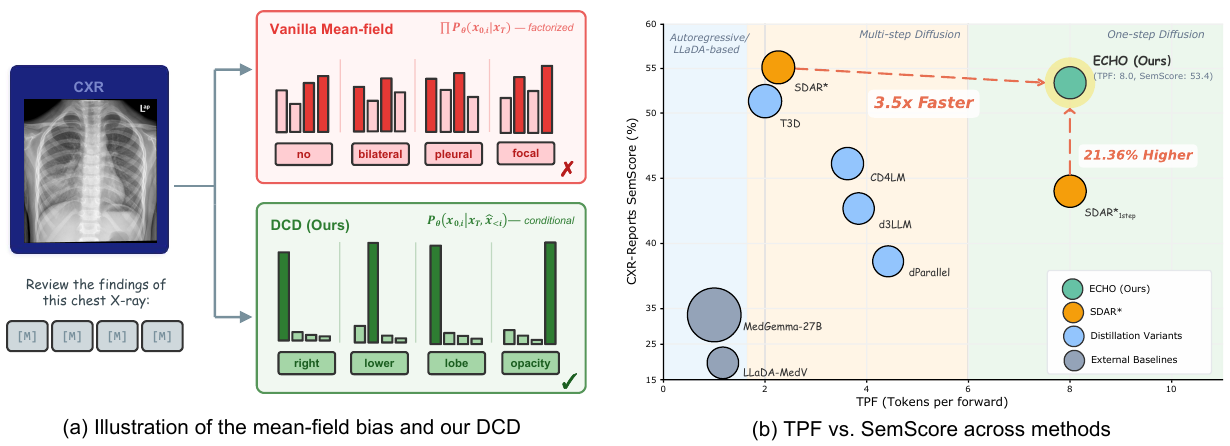}
    \caption{\textbf{Motivation and performance of ECHO.}
    \textbf{(a)} Decoding all tokens simultaneously in one step produces incoherent outputs, as standard diffusion models predict each position independently.
    Our Direct Conditional Distillation (DCD) distills from a non-factorized target, yielding coherent one-step-per-block outputs.
    \textbf{(b)} Compared to both autoregressive and diffusion-based baselines, \ours achieves a favorable trade-off between generation quality (SemScore) and decoding throughput (tokens per forward pass). Here, SDAR* denotes SDAR~\cite{cheng2025sdar} implemented under the multimodal large language model setting.}
    \label{fig:teaser}
  \end{figure*}

In recent years, Vision-Language Models (VLMs)~\cite{liu2023visual,lillava,li2024llava,wang2024qwen2,bai2025qwen3,hurst2024gpt,team2023gemini}, where visual features are aligned with language instructions to enable complex cross-modal understanding, have demonstrated significant progress in many fields, notably medical image analysis ~\cite{li2023llava,saab2024capabilities,sellergren2025medgemma,tu2024towards,xu2025lingshu,jiang2025hulu,pan2025medvlm,chen2024towards,zhang2024generalist}. 
Within this field, automated chest X-ray report generation (CXR-RG)~\cite{chen2024vision,lee2025cxr,pellegrini2023radialog,thawakar2024xraygpt,yang2025medxchat,wu2025towards} has emerged as a critical application. As one of the most common clinical imaging exams, CXR’s high volume places a heavy diagnostic burden on radiologists, creating strong demand for high-throughput automated reporting systems to ease workloads. 
Despite the promising performance achieved, canonical autoregressive (AR) VLMs often suffer from high inference latency due to their sequential decoding mechanism, which becomes a primary bottleneck for producing reports rapidly and at scale.
Fortunately, thanks to parallel decoding capabilities, emerging diffusion-based Vision Language Models (dVLMs)~\cite{yu2025dimple,lilavida,cheng2025sdar,you2025llada} offer a highly promising route toward achieving fast report generation.

 Despite the promise of parallel decoding, dVLMs in practice necessitate multiple denoising steps to ensure output coherence. This requirement stems from the mean-field approximation underlying their token-factorized denoisers, which introduces a structural bias that scales monotonically with the noise ratio. Although compressing decoding into a single step theoretically maximizes throughput, it compels the model to predict all tokens simultaneously from a fully masked input, precisely the scenario where mean-field bias is most acute. As illustrated in Fig.~\ref{fig:teaser}, the resulting inter-token incoherence leads to substantial quality degradation, prompting the critical question: \emph{\textbf{Can we achieve the upper bound of decoding speed without compromising output fidelity?}}

To address this challenge, distillation emerges as a natural solution, seeking to transfer the quality of multi-step denoising into a student model that operates in fewer passes. Specifically, several dLLMs ~\cite{deschenaux2025sdtt,liang2026cd4lm,qian2026d3llm,zhang2026t3d,chen2025dparallel} implement this via self-distillation, which aligns token-wise predictions with teacher's ones from less corrupted states. However, these teacher targets remain factorized across positions, ignoring dependencies between concurrently predicted tokens. For few-step inference, this factorization remains tolerable, as progressive unmasking provides the inter-token context that token-independent targets fail to capture. In contrast, one-step decoding lacks such a corrective mechanism, allowing the full mean-field bias to resurface unchecked. Therefore, enabling reliable one-step decoding requires a fundamentally different training objective, one that is unfactorized and directly encodes the joint dependencies among tokens predicted in parallel. To achieve this goal, we propose a new Direct Conditional Distillation (DCD), which constructs supervision from the teacher's on-policy remasking trajectory by conditioning each target on committed high-confidence context. The constructed supervision can encode inter-token dependencies into the training signal and enable stable one-step-per-block parallel inference.

Building upon DCD, we present \textbf{ECHO} (\textbf{E}fficient \textbf{Ch}est X-ray Report Generation with \textbf{O}ne-step Block Diffusion), a new foundational dVLM for CXR report generation that achieves both clinical accuracy and one-step inference efficiency. More concretely, we first establish an enhanced AR-based CXR-RG VLM by training on our curated data, where clinical findings and symptoms are explicitly and comprehensively annotated. Based on this, we propose Response-Asymmetric Diffusion (RAD) adaptation to efficiently convert the AR model into a block diffusion decoding paradigm. Subsequently, we apply DCD to distill this multi-step model into a one-step counterpart.

Notably, compared with state-of-the-art autoregressive methods, \ours improves RaTE and SemScore by 64.33\% and 60.58\%, respectively, while achieving up to $8\times$ theoretical and $5.1\times$ practical inference speedup over autoregressive baselines. Our main contributions are summarized below:
\begin{itemize}[topsep=3pt, itemsep=2pt, parsep=0pt]
    \item We present \textbf{ECHO}, a novel dVLM for CXR report generation that delivers strong clinical accuracy through one-step-per-block parallel decoding, outperforming both pioneering autoregressive and diffusion-based models by a large margin across standard benchmarks.
    \item We propose a novel \textbf{Direct Conditional Distillation (DCD)}, to the best of our knowledge the first one-step distillation framework for discrete diffusion language models. By enabling one-step-per-block inference that reaches the upper bound of decoding speed, DCD achieves up to 390\% inference speedup over the corresponding multi-step baseline at block size $L{=}8$ with only marginal quality degradation.
    \item We design \textbf{Response-Asymmetric Diffusion (RAD)} adaptation, which further reduces theoretical training FLOPs by 72.3\%, translating to a $3.61\times$ speedup in training efficiency.
\end{itemize}

\section{Related Works}
\label{related}

\subsection{Multimodal Medical Foundation Models}
Large-scale vision-language models (VLMs) ~\cite{liu2023visual,lillava,li2024llava,wang2024qwen2,bai2025qwen3,hurst2024gpt,team2023gemini} have established strong multimodal understanding by aligning visual representations with language instructions, achieving remarkable performance across diverse visual reasoning tasks.
Motivated by this progress, researchers have adapted VLMs to the medical domain ~\cite{li2023llava,saab2024capabilities,sellergren2025medgemma,tu2024towards,xu2025lingshu,jiang2025hulu,pan2025medvlm,chen2024towards,zhang2024generalist}, where precise alignment between clinical images and textual descriptions is essential.
Within medical imaging, automated CXR report generation has emerged as a central task ~\cite{chen2024vision,lee2025cxr,pellegrini2023radialog,thawakar2024xraygpt,yang2025medxchat,wu2025towards}, where existing methods predominantly follow an autoregressive paradigm and achieve strong clinical accuracy but at the cost of sequential decoding throughput, motivating the need for faster generation paradigms.

\subsection{Diffusion Language Model}
Discrete diffusion language models (dLLMs) represent one such paradigm, enabling parallel token prediction in place of sequential decoding.
Specifically, dLLMs ~\cite{austin2021structured,nielarge,ye2025dream,shi2024simplified,bie2025llada2} define a forward process that progressively masks tokens and a learned reverse process that recovers them over a discrete token vocabulary.
Building on this foundation, dVLMs ~\cite{yu2025dimple,lilavida,cheng2025sdar,you2025llada} extend the framework to vision-language tasks through a two-stage process of visual encoder alignment followed by instruction fine-tuning, enabling image-conditioned parallel generation.
Block diffusion ~\cite{arriolablock,fathi2025unifying,han2023ssd} further refines these models by adopting a semi-autoregressive decoding scheme that generates outputs block by block in causal order, naturally supporting variable-length sequence generation.
A further line of work specifically addresses the high training cost of building such models from scratch by directly adapting pretrained autoregressive models into block diffusion models ~\cite{cheng2025sdar,wu2025fast2,gongscaling,rnd1_2025,bie2025llada2}, offering a practical and scalable alternative.
Collectively, these advances have improved both the training scalability and inference throughput of dLLMs and dVLMs.

\subsection{Acceleration of dLLMs}
As dLLMs and dVLMs continue to mature, further accelerating their inference has attracted growing research attention.
This has been pursued from two complementary directions: inference-time optimization techniques ~\cite{wu2025fast,hu2025accelerating,liu2025dllm,christopher2025speculative,li2025diffuspec,wang2025diffusion} that reduce per-step or per-token computation overhead, and training-time distillation~\cite{deschenaux2025sdtt,liang2026cd4lm,qian2026d3llm,zhang2026t3d,chen2025dparallel} that compresses the multi-step denoising process into fewer steps.
Along the distillation direction, methods such as SDTT ~\cite{deschenaux2025sdtt} and dParallel ~\cite{chen2025dparallel} progressively align predictions at higher noise levels with those at lower ones, using either cross-entropy loss on self-generated trajectories or KL divergence between predicted distributions.
Most recently, T3D ~\cite{zhang2026t3d} takes a distinct angle, employing a DPO-style optimization objective to directly penalize mean-field bias during training rather than matching noise-level predictions.
Despite these advances, all existing methods retain token-factorized prediction targets and therefore still require multiple denoising steps to produce coherent outputs, leaving dLLM and dVLM throughput far below its theoretical upper bound.
\ours addresses this gap with Direct Conditional Distillation (DCD), a distillation framework for dLLMs and dVLMs that supports one-step denoising, applied here to block diffusion.

\section{Preliminaries} \label{sec:mean_field}

To motivate our approach, we formalize the dLLM framework and analyze how its token-factorized parameterization gives rise to the parallelism bottleneck identified in Section~\ref{sec:intro}.
Let $\mathcal{V}$ denote the vocabulary and $L$ the sequence length. A token sequence is $\mathbf{x}_0 = (x_{0,1}, \dots, x_{0,L}) \in \mathcal{V}^L$. Discrete diffusion language models (dLLMs)~\cite{ye2025dream,shi2024simplified,sahoo2024simple} define a forward process $q(\mathbf{x}_{1:T} \mid \mathbf{x}_0)$ that progressively masks tokens, and a reverse process $p_\theta(\mathbf{x}_{0:T-1} \mid \mathbf{x}_T)$ that recovers them.

\noindent\textbf{Mean-field parameterization.}
Since the joint posterior $p(\mathbf{x}_0 \mid \mathbf{x}_t, t)$ over $\mathcal{V}^L$ requires exponentially many parameters, existing dLLMs~\cite{sahoo2024simple,arriolablock,nielarge} universally adopt a token-factorized (mean-field) approximation~\cite{austin2021structured,xuenergy,yooredi,zhang2025variational}:
\begin{equation}
p_\theta(\mathbf{x}_0 \mid \mathbf{x}_t, t) \;=\; \prod_{i=1}^{L}\, p_\theta(x_{0,i} \mid \mathbf{x}_t, t),
\label{eq:mean_field}
\end{equation}
trained by minimizing the per-token denoising objective:
\begin{equation}
\mathcal{L}_{\mathrm{MF}}(\theta) \;=\; \mathbb{E}_{\mathbf{x}_0,\, t,\, \mathbf{x}_t} \!\left[ -\sum_{i=1}^{L} \log\, p_\theta(x_{0,i} \mid \mathbf{x}_t, t) \right].
\label{eq:mf_loss}
\end{equation}
As Eq.~\eqref{eq:mf_loss} decomposes into $L$ independent per-position cross-entropies, the global optimum is attained when each factor independently matches the true conditional marginal, i.e.\ $p_\theta^{\star}(x_{0,i} \mid \mathbf{x}_t, t) = q(x_{0,i} \mid \mathbf{x}_t, t)$. The resulting optimal factorized joint $p_{\mathrm{MF}}^{\star} = \prod_{i} q(x_{0,i} \mid \mathbf{x}_t, t)$ aligns each position's marginal with the true posterior, but cannot capture the cross-positional correlations present in $q(\mathbf{x}_0 \mid \mathbf{x}_t, t)$. We measure this gap by the mean-field bias:
\begin{equation}
\epsilon_{\mathrm{MF}}(\mathbf{x}_t, t) \;\triangleq\; \mathrm{KL}\!\bigl( q(\mathbf{x}_0 \mid \mathbf{x}_t, t) \,\big\|\, p_{\mathrm{MF}}^{\star}(\mathbf{x}_0 \mid \mathbf{x}_t, t) \bigr),
\label{eq:mf_bias}
\end{equation}
which quantifies the irreducible joint dependence structure across token positions that no factorized distribution can represent.

\noindent\textbf{Why multi-step sampling mitigates the bias.}
The expected mean-field bias $\bar{\epsilon}_{\mathrm{MF}}(t) \triangleq \mathbb{E}_{\mathbf{x}_t}[\epsilon_{\mathrm{MF}}(\mathbf{x}_t, t)]$ is governed by the corruption level: when $\mathbf{x}_t$ has few masked tokens, the remaining unknowns are sparsely distributed and largely conditionally independent, so $\bar{\epsilon}_{\mathrm{MF}}(t)$ is small; when all tokens are masked ($t = T$), the posterior exhibits strong cross-positional dependence and $\bar{\epsilon}_{\mathrm{MF}}(T)$ reaches its maximum. In general, $\bar{\epsilon}_{\mathrm{MF}}(t)$ grows monotonically with the number of masked positions.~\cite{yooredi} Multi-step reverse sampling systematically exploits this monotonicity. At each reverse step $s$, tokens decoded with high confidence from $p_\theta(\cdot \mid \mathbf{x}_s, s)$ are committed as observed context, reducing the number of masked positions from $m_s$ to $m_{s-1} < m_s$. Consequently, the model at step $s-1$ operates under a strictly reduced corruption level and, accordingly, a lower expected bias $\bar{\epsilon}_{\mathrm{MF}}(s-1) \leq \bar{\epsilon}_{\mathrm{MF}}(s)$. By iterating this process, the multi-step chain progressively shifts decoding from the high-bias fully-masked regime toward a low-bias, nearly-observed state, thereby recovering the inter-token dependence structure that Eq.~\eqref{eq:mean_field} otherwise discards.

\section{Methodology}

\subsection{Overview}

\begin{figure*}[t]
    \centering
    \includegraphics[width=\textwidth]{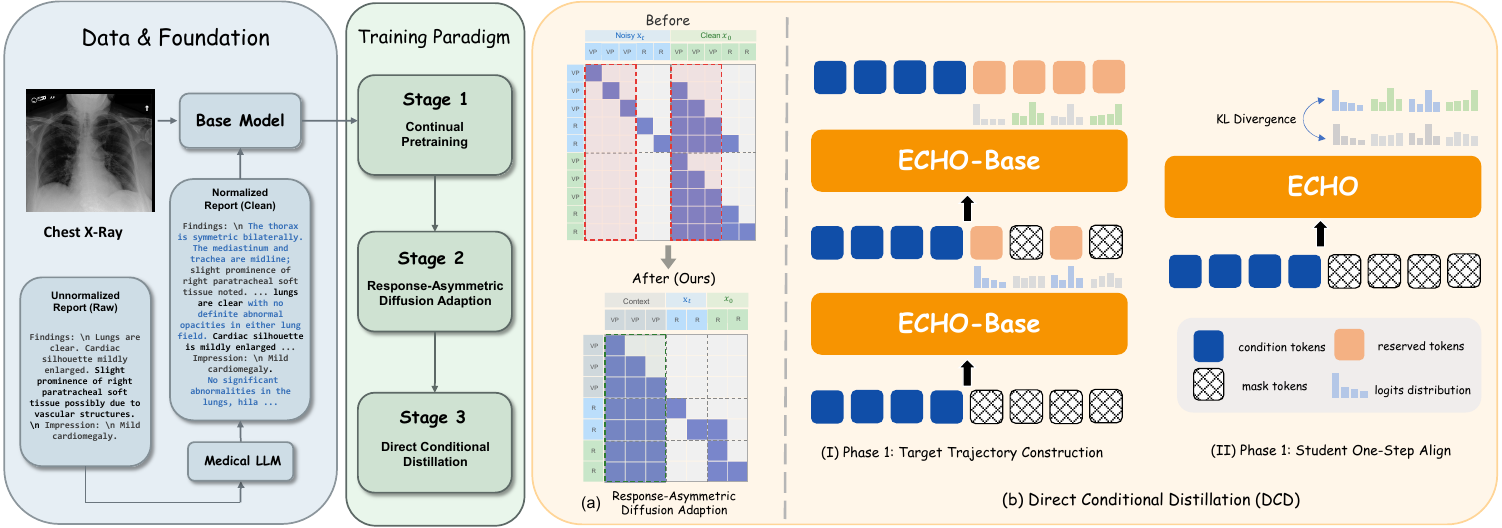}
    \caption{\textbf{Overview of the ECHO training pipeline.}
    ECHO is built in three successive stages: continued pre-training (Stage~1) produces \ar; Response-Asymmetric Diffusion adaptation (Stage~2) converts it into the block diffusion model \base; and Direct Conditional Distillation (Stage~3) distills \base into the final one-step-per-block model \ours.
    \textbf{(a)} RAD duplicates only the response portion of the training sequence, avoiding the redundant duplication of long vision token contexts required by prior two-stage conversion methods.
    \textbf{(b)} DCD proceeds in two phases: the teacher's confidence-heuristic remasking trajectory is collected to form a joint, non-factorized supervision target, which is then used to align the student's single-step prediction via KL divergence.} 
    \label{fig:pipeline}
\end{figure*}

As illustrated in Fig.~\ref{fig:pipeline}, the training pipeline of \ours comprises three successive stages.
In Stage~1, we perform continued pre-training (CPT) on Lingshu-7B~\cite{xu2025lingshu} using a curated CXR corpus.
This produces \ar, an autoregressive vision-language model specialized in radiology report generation.
Subsequently, in Stage~2, we propose Response-Asymmetric Diffusion (RAD) adaptation, which converts \ar into \base, a block diffusion decoding model that retains the domain knowledge of \ar, achieving an initial inference speedup.
Lastly, in Stage~3, to further push decoding speed toward its theoretical upper bound, we apply Direct Conditional Distillation (DCD) to train \base toward one-step-per-block decoding, yielding the final \ours model.

\subsection{From ECHO-AR to ECHO}
\label{sec:RAD}
\begin{figure*}[t]
    \centering
    \includegraphics[width=\textwidth]{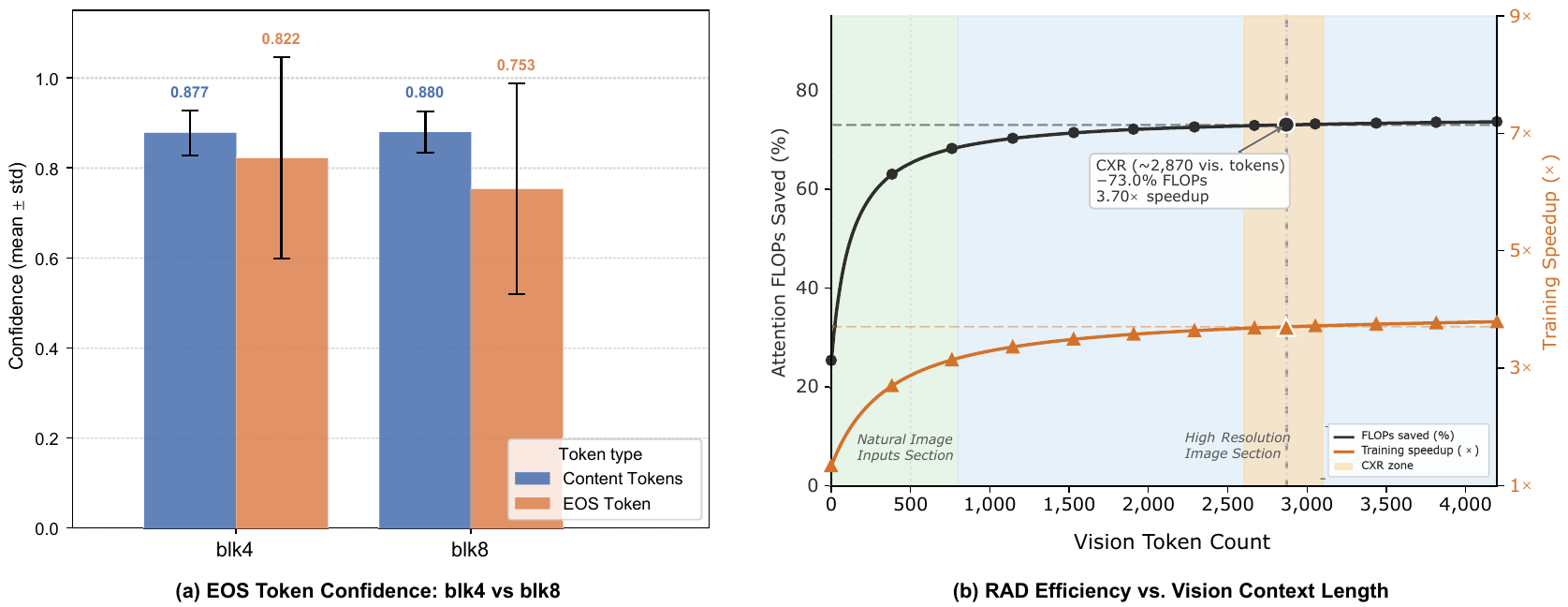}
    \caption{\textbf{(a)} Confidence analysis of \texttt{<eos>} tokens vs.\ content tokens under block sizes blk4 and blk8 in standard multi-step inference.\textbf{(b)} RAD attention FLOPs saved (\%) and training speedup ($\times$) as a function of vision token count.}
    \label{fig:rad_eos}
\end{figure*}

\noindent\textbf{Response-Asymmetric Diffusion Adaptation.}
While \ar achieves high-quality report generation, its token-by-token decoding mechanism limits inference throughput.
To address this, our first objective is to convert \ar into \base, a block diffusion model that preserves the domain knowledge acquired during pretraining.
Prior work~\cite{cheng2025sdar} has implemented this through a two-stage adaptation, i.e., a pretraining phase followed by SFT.
In these methods, the entire training sequence, covering vision, instruction, and response tokens, is duplicated across both stages to construct block-causal teacher-forcing targets.
For CXR report generation, where vision token sequences are substantially long, this full-sequence duplication incurs prohibitive training costs.
To overcome this inefficiency, we propose Response-Asymmetric Diffusion (RAD), which achieves this adaptation through a single SFT stage.
As illustrated in Fig.~\ref{fig:pipeline}, RAD duplicates only the response portion of the training sequence, and constructs a block attention mask such that each noisy response block attends to all vision and instruction tokens as well as all previously decoded blocks.
As illustrated in Fig.~\ref{fig:rad_eos}(b), this asymmetric design eliminates the redundant duplication of long vision token contexts, substantially reducing training FLOPs while consolidating the originally two-stage conversion into one.
Furthermore, we conduct a series of experiments (Sec.~\ref{exp:rad_data_analysis}) to investigate the relationship between training data volume, model quality, and inference speed during the RAD conversion.
We find that \base attains performance comparable to \ar using only a small fraction of the original training data, demonstrating the high knowledge-transfer efficiency of RAD.

\noindent\textbf{Direct Conditional Distillation.}
With \base established, our next objective is to convert it into a model capable of one-step-per-block decoding, fully maximizing inference throughput.
However, as discussed in Sec.~\ref{sec:mean_field}, discrete diffusion language models inherently rely on multi-step denoising to progressively reduce the mean-field bias, and our dVLM setting is no exception.
This calls for a training target that is itself non-factorized: by aligning the student to such a target, the inter-token dependencies accumulated across the multi-step denoising trajectory can be captured in a single forward pass.
To this end, we propose Direct Conditional Distillation (DCD), which constructs this non-factorized target by collecting and stitching the high-confidence token distributions at each step of the teacher's multi-step denoising process, forming a joint supervision signal that encodes cross-token dependencies.

\begin{algorithm}[t]
    \caption{Direct Conditional Distillation (DCD)}
    \label{alg:dcd}
    \SetKwInOut{Input}{Input}
    \SetKwInOut{Output}{Output}

    \Input{Context $\mathbf{x}_{\text{ctx}}$, teacher $\theta$, student $\phi$, $N$ blocks, block length $L$, sampling confidence threshold $\tau$}
    \Output{Updated student $\phi$}

    \tcp{Phase 1: On-policy Teacher Trajectory Collection}
    \For{$n = 1, \dots, N$}{
        Initialize $\mathcal{B} \leftarrow \{1,\dots,L\}$,\;
        $\mathbf{x}_{\text{curr}} \leftarrow (\mathbf{x}_{\text{ctx}},\,\hat{\mathbf{b}}_{1:n-1})$\;
        \While{$\mathcal{B} \neq \emptyset$}{
            Query teacher $p_{\theta}(x_i \mid \mathbf{x}_{\text{curr}})$ for all $i \in \mathcal{B}$\;
            $\mathcal{U} \leftarrow \{i \in \mathcal{B} \mid c_i \ge \tau\}$;\;
            \lIf{$\mathcal{U} = \emptyset$}{$\mathcal{U} \leftarrow \{\arg\max_{i \in \mathcal{B}}\, c_i\}$}
            $\mathcal{P}_{\text{tch}}^{(n)}[i] \leftarrow p_{\theta}(x_i \mid \mathbf{x}_{\text{curr}})$,\;
            $\hat{x}_i \leftarrow \arg\max\, p_{\theta}(x_i \mid \mathbf{x}_{\text{curr}})$ for all $i \in \mathcal{U}$\;
            $\mathbf{x}_{\text{curr}} \leftarrow \mathbf{x}_{\text{curr}} \cup \{\hat{x}_i\}_{i \in \mathcal{U}}$,\;
            $\mathcal{B} \leftarrow \mathcal{B} \setminus \mathcal{U}$\;
        }
        $\hat{\mathbf{b}}_n \leftarrow (\hat{x}_1, \dots, \hat{x}_L)$\;
    }

    \tcp{Phase 2: Student One-Step Alignment (cf.\ Sec.~\ref{sec:RAD} and Fig.~\ref{fig:pipeline})}
    Construct RAD training sequence $\mathbf{x}_{\text{train}}$ using $(\mathbf{x}_{\text{ctx}},\,\hat{\mathbf{b}}_{1:N})$\;
    $\mathcal{L}_{\text{DCD}} \leftarrow \displaystyle\sum_{n,i} D_{\text{KL}}\!\bigl(\mathcal{P}_{\text{tch}}^{(n)}[i] \;\big\|\; p_{\phi}(b_{n,i} \mid \mathbf{x}_{\text{train}})\bigr)$\;
    Update $\phi \leftarrow \text{Optimizer}(\phi,\, \nabla_{\phi}\mathcal{L}_{\text{DCD}})$\;
\end{algorithm}

The full self-distillation procedure is detailed in Algorithm~\ref{alg:dcd} and consists of the following two phases, applied iteratively during training.

\noindent\textbf{Phase 1: On-policy Teacher Trajectory Collection.}
The teacher trajectory is the trajectory sampled by its multi-step denoising process. To obtain this, we first run \base through a confidence-heuristic denoising process, where tokens are progressively unmasked in order of their prediction confidence $c_i \triangleq \max_{v} p_\theta(x_i{=}v \mid \mathbf{x}_{\text{curr}})$.
Specifically, at each step of a block's denoising, we record the predicted distribution and the committed token label for every position as it is unmasked.
The collected token labels form the pseudo labels $\hat{\mathbf{b}}_n$, which are used to construct the training sequence for the subsequent student alignment phase.
The collected distributions are then stitched together into the joint target $\mathcal{P}_{\text{tch}}^{(n)}$ for the $n$-th block, as the distillation target in Phase~2.

\noindent\textbf{Phase 2: Student One-Step Alignment.}
With the pseudo-labels and stitched joint distribution collected in Phase~1, the next step is to align this distribution with the student's one-step prediction.
Concretely, we first use the pseudo-labels to construct a block teacher-forcing training sequence following the RAD scheme (Sec.~\ref{sec:RAD}), ensuring that the student makes its prediction under exactly the same conditions as the teacher.
Letting $b_{n,i}$ denote the token at position $i$ of block $n$ and $\mathcal{Q}_{\phi}^{(n)} \triangleq \bigl\{p_\phi(b_{n,i} \mid \mathbf{x}_{\text{train}})\bigr\}_{i=1}^{L}$ the student's one-step predicted distribution over that block, we minimize the forward Kullback–Leibler (KL) divergence $D_{\mathrm{KL}}\!\left(\mathcal{P}_{\text{tch}}^{(n)} \,\|\, \mathcal{Q}_{\phi}^{(n)}\right)$ for each block and aggregate the loss over all blocks.
Intuitively, a token that is unmasked at a later step within a block requires stronger conditioning to reach a high-confidence prediction, indicating that its position is subject to more severe mean-field bias.
Therefore, we introduce a token reweighting scheme that assigns each position a weight proportional to the step at which it is unmasked.
Under this scheme, positions with larger mean-field bias receive stronger supervision, while tokens committed at early steps serve as a regularization signal for the alignment process.

After DCD, we can generate each block with only one step, achieving fast decoding speed without compromising clinical accuracy.

\subsection{Hallucination Mitigation}
Hallucination is a critical concern in CXR report generation, where inaccurate descriptions of clinical findings can directly undermine diagnostic reliability.
During the development of our vision-language model, we identify two predominant hallucination patterns in medical dLLMs: inaccurate identification of symptoms, and degenerate repetition loops that fail to terminate.

For inaccurate symptom identification, we attribute this pattern to an implicit negative bias inherent in the training data.
In routine clinical practice, radiologists follow a ``reporting by exception'' convention, describing only abnormal findings while omitting normal structures or dismissing them with a single catch-all phrase such as ``no other significant abnormality.''
Consequently, for the majority of anatomical regions that are clinically unremarkable, their normal status is absent from the report rather than explicitly stated as negative.
This systematic omission leaves the model without explicit negative evidence during training, leaving its conditional distribution under-constrained.
At inference time, this under-constraint manifests as two complementary failure modes: false-positive hallucinations, where the model fabricates findings for normal regions, and false-negative omissions, where genuine abnormalities are overlooked.
To address this, we propose a data normalization paradigm tailored for CXR report generation.
Specifically, we reformulate every training report so that each predefined anatomical region receives an explicit annotation, either a positive finding or a negative assertion.
This ensures unambiguous supervision at every position, directly eliminating the implicit omission bias described above.
Our experiments in Sec.~\ref{exp:report_normalization_benefits} show that the hallucination reduction brought by data normalization consistently benefits both \ar and the final \ours model.

Beyond inaccurate symptom identification, \ours also suffers from degenerate repetition loops that fail to terminate.
To understand the root cause, we examine \base's prediction confidence of content tokens and \texttt{<eos>} tokens under standard confidence-heuristic inference.
As shown in Fig.~\ref{fig:rad_eos}(a), \texttt{<eos>} tokens exhibit systematically lower mean confidence and substantially greater variance compared to content tokens.
Using such a flat and unstable distribution as the distillation target for the \texttt{<eos>} token makes it difficult for the one-step model to terminate generation with high confidence.
This problem is further exacerbated as the block size grows: as shown in Fig.~\ref{fig:rad_eos}(a), increasing the block size leaves content token confidence nearly unaffected while causing a notable decline in \texttt{<eos>} confidence.
To address this, we apply an additional cross-entropy loss on the \texttt{<eos>} token during distillation, explicitly pulling its predicted distribution toward a sharp, high-confidence one-hot target.
Together, the two hallucination mitigation strategies consistently reduce generation errors in \ours, contributing to improved clinical accuracy.

\begin{figure*}[t]
    \centering
    \includegraphics[width=\textwidth]{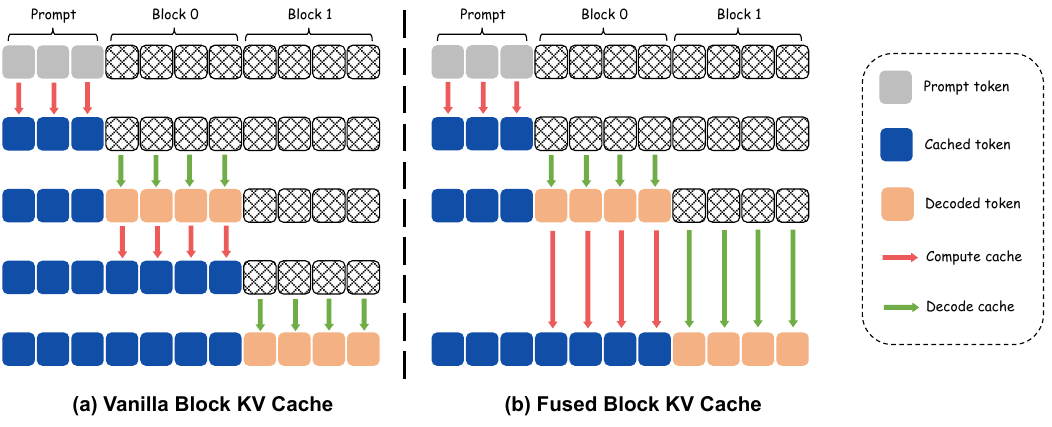}
    \caption{\textbf{(a)} Vanilla block KV cache: after all tokens of a block are committed, a dedicated forward pass is performed to update the KV cache.
    \textbf{(b)} Fused block KV cache: the KV cache update for the preceding block is fused into the current block's denoising forward, eliminating the dedicated KV update pass.}
    \label{fig:kv_cache}
\end{figure*}

\subsection{Inference Paradigm}
To further maximize the inference throughput of \ours, we optimize the inference paradigm used in dLLMs.
A common technique in semi-autoregressive decoding is block KV cache~\cite{wu2025fast}, which caches the key-value states of previously decoded blocks to avoid redundant recomputation.
As illustrated in Fig.~\ref{fig:kv_cache}(a), after all tokens of a block are committed, a dedicated forward pass is performed solely to update the KV cache with the newly decoded block.
This additional forward pass incurs acceptable inference time overhead in multi-step denoising, but for our one-step model it doubles the total number of forward passes for each block.
To eliminate this overhead, we propose \noindent\textbf{Fused Block KV Cache.}
As illustrated in Fig.~\ref{fig:kv_cache}(b), after a block is decoded, we defer its KV cache update and fuse it into the next block's denoising forward.
In this fused forward, the model simultaneously computes and caches the key-value states for the preceding decoded block while denoising the current block's fully-masked tokens, removing the dedicated KV update pass entirely.
As proven in Sec.~\ref{sec:proof_fused_kv}, Fused Block KV Cache introduces no additional FLOPs while halving the number of forward passes, directly reducing inference latency.
\section{Experiments}
\label{sec:experiment}

In this section, we present experimental results to assess the effectiveness of \ours for fast and reliable CXR report generation.
We compare \ours against state-of-the-art autoregressive baselines on report quality, and against diffusion-based distillation baselines on distillation efficiency.
We then ablate the individual components of DCD, and analyze the effect of training data volume during RAD adaptation and the impact of report normalization across all training stages.

\subsection{Experimental Setups}
\subsubsection{Dataset}
We conduct experiments on a unified CXR report corpus built from multiple public datasets, including MIMIC-CXR~\cite{johnson2019mimic}, CheXpert-Plus~\cite{irvin2019chexpert}, ReXGradient~\cite{zhang2025rexgradient}, and IU-Xray~\cite{demner2016preparing}.\\
For training, we apply a standardized cleaning and normalization pipeline across all source datasets to construct a bilingual training set, with a small subset of LLaVA-ReCap-558K~\cite{li2024llava} included to mitigate catastrophic forgetting.
For evaluation, we sample 2,000 English and 2,000 Chinese reports from each of normalized MIMIC-CXR, CheXpert-Plus, and ReXGradient, ensuring balanced coverage across datasets and languages.
Full preprocessing details and data statistics are provided in Appendix.~\ref{sec:appendix-data-details}.

\subsubsection{Models}
All \ours models are initialized from Lingshu-7B~\cite{xu2025lingshu}, an autoregressive VLM pretrained on large-scale medical corpora, which provides a strong domain-specific knowledge and vision alignment for CXR report generation.

\subsubsection{Implementation Details}
For RAD, we use the same data as the SFT stage in continued pre-training.
The vision encoder and projector weights of \ar are frozen throughout, and only the LLM backbone is trained.
For DCD, we randomly sample 30{,}000 examples from the RAD training set, drawn proportionally across datasets.
During teacher trajectory collection, samples that produce degenerate repetition loops are automatically discarded.

\subsubsection{Evaluation Metrics}
To provide a comprehensive assessment of the generated chest X-ray reports, we evaluate our model across four complementary dimensions: linguistic quality, clinical fidelity, structural stability, and decoding efficiency. These metrics enable a rigorous comparison between \ours and existing state-of-the-art (SOTA) models while specifically addressing the unique challenges of diffusion-based generation.
\begin{itemize}[topsep=3pt, itemsep=2pt, parsep=0pt]
    \item \textbf{Linguistic Quality Metrics (LQM).} We use standard natural language generation (NLG) metrics, including ROUGE-L~\cite{lin2004rouge} and CIDEr~\cite{vedantam2015cider}, to measure lexical overlap and fluency between generated and reference reports, enabling direct comparison with prior methods.

    \item \textbf{Clinical Fidelity Metrics (CFM).} Linguistic fluency does not guarantee accurate symptom identification, so we use RaTEScore~\cite{zhao2024ratescore} and SemScore~\cite{smit2020combining} to assess clinical content fidelity. Since our evaluation set consists of normalized reports, a fair comparison across models requires focusing solely on symptom identification accuracy rather than reporting style. To this end, we only adopt the positive-finding score of RaTEScore, and SemScore is insensitive to negative findings by design.

    \item \textbf{Structural Stability Metrics (SSM).} To assess generation stability specific to discrete diffusion models, we report Perplexity (PPL) computed with Qwen3-1.7B as the judge model. Lower PPL indicates more stable and fluent generation.

    \item \textbf{Efficiency Metrics (EM).} To evaluate decoding efficiency, we report two complementary speed metrics. TPF (Tokens Per Forward Pass) measures theoretical throughput as the number of decoded tokens per forward pass, normalized relative to the AR baseline ($\times 1.0$). TPS (Tokens Per Second) measures practical decoding throughput as the number of tokens generated per second during inference.
\end{itemize}

\begin{table}[t]
  \centering
  \caption{\textbf{Performance comparison with state-of-the-art models.} We report average metrics across the MIMIC-CXR, CheXpert-Plus, and ReXGradient test sets. \textbf{Bold} values indicate the best performance. Colored superscript percentages indicate relative quality degradation with respect to \ar. Please refer to Appendix.~\ref{sec:appendix-detailed-results} for detailed results on individual datasets.}
  \label{tab:performance_comparison}
  \setlength{\tabcolsep}{3.5pt}
  \small
  \resizebox{\linewidth}{!}{%
\begin{tabular}{l cccc cccc cccc cc}
    \toprule
    \multirow{2}{*}{\textbf{Methods}} & \multicolumn{4}{c}{\textbf{CheXpert-Plus}} & \multicolumn{4}{c}{\textbf{ReXGradient}} &  \multicolumn{4}{c}{\textbf{MIMIC-CXR}} & \multicolumn{2}{c}{\textbf{Speed}} \\
    \cmidrule(lr){2-5} \cmidrule(lr){6-9} \cmidrule(lr){10-13} \cmidrule(lr){14-15}
    & ROUGE-L & CIDEr & RateScore & SemScore & ROUGE-L & CIDEr & RateScore & SemScore & ROUGE-L & CIDEr & RateScore & SemScore & TPF & TPS \\
    \midrule

    \multicolumn{15}{c}{\textbf{Proprietary general models}} \\
    \midrule
    Gemini3-Pro~\cite{deepmind_gemini3pro_modelcard_2025}     & 26.95 & 1.53 & 40.86 & 29.72 & 32.10 & 1.81 & 33.52 & 39.02 & 27.08 & 1.55 & 41.42 & 30.45 & $\times$1.0 & -- \\
    Qwen3-Max~\cite{bai2025qwen3}       & 27.19 & 1.53 & 41.63 & 29.24 & 30.50 & 1.72 & 39.12 & 38.13 & 26.86 & 1.51 & 40.78 & 27.32 & $\times$1.0 & -- \\
    \midrule

    \multicolumn{15}{c}{\textbf{Autoregressive Medical Models}} \\
    \midrule
    LLaVA-Med~\cite{li2023llava}   & 6.92 & 0.65 & 10.11 & 11.39 & 7.36 & 0.72 & 10.01 & 17.04 & 6.69 & 0.65 & 10.07 & 9.22 & $\times$1.0 & 36.33 \\
    Lingshu-7B~\cite{xu2025lingshu}      & 21.95 & 1.12 & 31.34 & 27.54 & 23.63 & 1.25 & 20.26 & 34.82 & 22.24 & 1.16 & 32.79 & 26.94 & $\times$1.0 & 53.70 \\
    Hulu-Med-7B~\cite{jiang2025hulu}     & 22.38 & 1.15 & 26.25 & 23.43 & 21.66 & 1.14 & 22.77 & 23.56 & 22.26 & 1.18 & 22.89 & 22.30 & $\times$1.0 & 38.78 \\
    MedGemma-27B~\cite{sellergren2025medgemma}    & 20.30 & 0.79 & 38.83 & 31.79 & 23.41 & 0.91 & 26.00 & 37.74 & 20.80 & 0.82 & 38.48 & 30.24 & $\times$1.0 & 16.78 \\
    Lingshu-32B~\cite{xu2025lingshu}     & 20.40 & 1.06 & 30.92 & 24.32 & 22.10 & 1.20 & 20.33 & 31.49 & 20.73 & 1.10 & 32.40 & 24.14 & $\times$1.0 & 23.60 \\
    Hulu-Med-32B~\cite{jiang2025hulu}    & 19.46 & 0.96 & 27.17 & 22.07 & 22.31 & 1.11 & 23.01 & 26.67 & 17.96 & 0.99 & 24.54 & 18.23 & $\times$1.0 & 15.67 \\
    \midrule
    \multicolumn{15}{c}{\textbf{Diffusion Medical Methods}} \\
    \midrule
    LLaDA-MedV~\cite{dong2025llada}      & 7.69 & 0.22 & 26.72 & 17.76 & 9.72 & 0.23 & 18.19 & 22.09 & 9.60 & 0.23 & 27.47 & 17.26 & $\times$1.30 & 12.86 \\
    \ar & 56.89 & 4.47 & 59.18 & 52.94 & 67.07 & 5.68 & 64.39 & 66.60 & 54.55 & 4.32 & 56.58 & 46.70 & $\times$1.0 & 53.70 \\
    \baseblkfour  & 56.90 & 4.25 & 59.12 & 52.64& 66.46 & 5.64& 63.02& 66.68& 54.45 & 4.16 & 55.85 & 46.10 & $\times$1.89 & 48.86 \\
    \baseblkeight & 55.97 & 4.19 & 58.39 & 50.13 & 65.62 & 5.47 & 62.48 & 64.90 & 53.52 & 3.98 & 54.96 & 44.48 & $\times$2.10 & 52.76 \\
    CD4LM~\cite{liang2026cd4lm}   & 49.21\smash{\rlap{$^{\textcolor{red}{14\%\downarrow}}$}} & 3.38\smash{\rlap{$^{\textcolor{red}{24\%\downarrow}}$}} & 55.90\smash{\rlap{$^{\textcolor{red}{6\%\downarrow}}$}} & 43.86\smash{\rlap{$^{\textcolor{red}{17\%\downarrow}}$}} & 58.73\smash{\rlap{$^{\textcolor{red}{12\%\downarrow}}$}} & 4.44\smash{\rlap{$^{\textcolor{red}{22\%\downarrow}}$}} & 58.36\smash{\rlap{$^{\textcolor{red}{9\%\downarrow}}$}} & 55.62\smash{\rlap{$^{\textcolor{red}{17\%\downarrow}}$}} & 46.21\smash{\rlap{$^{\textcolor{red}{15\%\downarrow}}$}} & 3.13\smash{\rlap{$^{\textcolor{red}{28\%\downarrow}}$}} & 52.11\smash{\rlap{$^{\textcolor{red}{8\%\downarrow}}$}} & 38.76\smash{\rlap{$^{\textcolor{red}{17\%\downarrow}}$}} & $\times$3.61 & 98.49 \\
    d3LLM~\cite{qian2026d3llm}  & 44.04\smash{\rlap{$^{\textcolor{red}{23\%\downarrow}}$}} & 2.92\smash{\rlap{$^{\textcolor{red}{35\%\downarrow}}$}} & 55.39\smash{\rlap{$^{\textcolor{red}{6\%\downarrow}}$}} & 40.35\smash{\rlap{$^{\textcolor{red}{24\%\downarrow}}$}} & 54.18\smash{\rlap{$^{\textcolor{red}{19\%\downarrow}}$}} & 4.17\smash{\rlap{$^{\textcolor{red}{27\%\downarrow}}$}} & 57.99\smash{\rlap{$^{\textcolor{red}{10\%\downarrow}}$}} & 53.50\smash{\rlap{$^{\textcolor{red}{20\%\downarrow}}$}} & 41.13\smash{\rlap{$^{\textcolor{red}{25\%\downarrow}}$}} & 2.65\smash{\rlap{$^{\textcolor{red}{39\%\downarrow}}$}} & 50.99\smash{\rlap{$^{\textcolor{red}{10\%\downarrow}}$}} & 34.20\smash{\rlap{$^{\textcolor{red}{27\%\downarrow}}$}} & $\times$3.84 & 101.64 \\
    dParallel~\cite{chen2025dparallel} & 42.22\smash{\rlap{$^{\textcolor{red}{26\%\downarrow}}$}} & 3.17\smash{\rlap{$^{\textcolor{red}{29\%\downarrow}}$}} & 44.65\smash{\rlap{$^{\textcolor{red}{25\%\downarrow}}$}} & 36.72\smash{\rlap{$^{\textcolor{red}{31\%\downarrow}}$}} & 48.73\smash{\rlap{$^{\textcolor{red}{27\%\downarrow}}$}} & 4.02\smash{\rlap{$^{\textcolor{red}{29\%\downarrow}}$}} & 45.50\smash{\rlap{$^{\textcolor{red}{29\%\downarrow}}$}} & 54.53\smash{\rlap{$^{\textcolor{red}{18\%\downarrow}}$}} & 39.98\smash{\rlap{$^{\textcolor{red}{27\%\downarrow}}$}} & 2.96\smash{\rlap{$^{\textcolor{red}{32\%\downarrow}}$}} & 40.47\smash{\rlap{$^{\textcolor{red}{29\%\downarrow}}$}} & 31.15\smash{\rlap{$^{\textcolor{red}{33\%\downarrow}}$}} & $\times$4.42 & 110.12 \\
    T3D~\cite{zhang2026t3d}    & 54.99\smash{\rlap{$^{\textcolor{red}{3\%\downarrow}}$}} & 4.06\smash{\rlap{$^{\textcolor{red}{9\%\downarrow}}$}} & 56.90\smash{\rlap{$^{\textcolor{red}{4\%\downarrow}}$}} & 49.86\smash{\rlap{$^{\textcolor{red}{6\%\downarrow}}$}} & 64.36\smash{\rlap{$^{\textcolor{red}{4\%\downarrow}}$}} & 5.23\smash{\rlap{$^{\textcolor{red}{8\%\downarrow}}$}} & 58.63\smash{\rlap{$^{\textcolor{red}{9\%\downarrow}}$}} & 61.65\smash{\rlap{$^{\textcolor{red}{7\%\downarrow}}$}} & 52.38\smash{\rlap{$^{\textcolor{red}{4\%\downarrow}}$}} & 3.85\smash{\rlap{$^{\textcolor{red}{11\%\downarrow}}$}} & 52.38\smash{\rlap{$^{\textcolor{red}{7\%\downarrow}}$}} & 42.51\smash{\rlap{$^{\textcolor{red}{9\%\downarrow}}$}} & $\times$2.00 & 60.25 \\
    \oursblkeight       & 55.21\smash{\rlap{$^{\textcolor{darkgreen}{3\%\downarrow}}$}} & 4.14\smash{\rlap{$^{\textcolor{darkgreen}{7\%\downarrow}}$}} & 56.85\smash{\rlap{$^{\textcolor{darkgreen}{4\%\downarrow}}$}} & 51.40\smash{\rlap{$^{\textcolor{darkgreen}{3\%\downarrow}}$}} & 65.13\smash{\rlap{$^{\textcolor{darkgreen}{3\%\downarrow}}$}} & 5.48\smash{\rlap{$^{\textcolor{darkgreen}{4\%\downarrow}}$}} & 59.92\smash{\rlap{$^{\textcolor{darkgreen}{7\%\downarrow}}$}} & 64.00\smash{\rlap{$^{\textcolor{darkgreen}{4\%\downarrow}}$}} & 53.07\smash{\rlap{$^{\textcolor{darkgreen}{3\%\downarrow}}$}} & 4.00\smash{\rlap{$^{\textcolor{darkgreen}{7\%\downarrow}}$}} & 52.97\smash{\rlap{$^{\textcolor{darkgreen}{6\%\downarrow}}$}} & 44.83\smash{\rlap{$^{\textcolor{darkgreen}{4\%\downarrow}}$}} & $\mathbf{\times8.00}$ & \textbf{274.21} \\
    \oursblkfour       & \textbf{56.14}\smash{\rlap{$^{\textcolor{darkgreen}{1\%\downarrow}}$}} & \textbf{4.20}\smash{\rlap{$^{\textcolor{darkgreen}{6\%\downarrow}}$}} & \textbf{57.40}\smash{\rlap{$^{\textcolor{darkgreen}{3\%\downarrow}}$}} & \textbf{49.57}\smash{\rlap{$^{\textcolor{darkgreen}{6\%\downarrow}}$}} & \textbf{66.39}\smash{\rlap{$^{\textcolor{darkgreen}{1\%\downarrow}}$}} & \textbf{5.54}\smash{\rlap{$^{\textcolor{darkgreen}{3\%\downarrow}}$}} & \textbf{62.18}\smash{\rlap{$^{\textcolor{darkgreen}{3\%\downarrow}}$}} & \textbf{66.28}\smash{\rlap{$^{\textcolor{darkgreen}{0\%\downarrow}}$}} & \textbf{54.12}\smash{\rlap{$^{\textcolor{darkgreen}{1\%\downarrow}}$}} & \textbf{4.05}\smash{\rlap{$^{\textcolor{darkgreen}{6\%\downarrow}}$}} & \textbf{53.95}\smash{\rlap{$^{\textcolor{darkgreen}{5\%\downarrow}}$}} & \textbf{45.57}\smash{\rlap{$^{\textcolor{darkgreen}{2\%\downarrow}}$}} & $\times$4.00 & 129.19 \\
    \bottomrule
  \end{tabular}%
  }
\end{table}

\subsection{Comparison with State-of-the-art Methods}
In Table~\ref{tab:performance_comparison}, to evaluate the effectiveness of \ours, we compare it against three categories of state-of-the-art models:
(1) General-purpose proprietary models, including Gemini3-Pro~\cite{deepmind_gemini3pro_modelcard_2025} and Qwen3VL-Max~\cite{bai2025qwen3}.
(2) Autoregressive (AR) medical VLMs, such as LLaVA-Med v1.5-7B~\cite{li2023llava}, Lingshu-7B/32B~\cite{xu2025lingshu}, MedGemma-27B~\cite{sellergren2025medgemma}, and Hulu-Med-7B/32B~\cite{jiang2025hulu}.  
(3) Diffusion-based medical models, including LLaDA-MedV-8B~\cite{dong2025llada} and several distilled variants ~\cite{liang2026cd4lm,qian2026d3llm,zhang2026t3d,chen2025dparallel}. These variants share the same base model \baseblkeight and distillation data, but with different distillation targets.

\noindent\textbf{Overall Performance.} \ours consistently achieves superior performance compared to both general and specialized models. Notably, it significantly outperforms larger medical VLMs like MedGemma-27B, ranging from 17\% to 40\% in CFM. Even when compared to the most powerful models like Gemini3-Pro and Qwen3VL-Max, \ours maintains a clear advantage in all medical clinical metrics. \\
\noindent\textbf{Efficiency of Distillation.} Since all diffusion-based methods compared here are ultimately derived from \ar, the percentage drops in Tab.~\ref{tab:performance_comparison} are reported relative to \ar, providing a unified basis for comparing distillation efficiency across methods and block size configurations.
As shown in the table, \oursblkeight achieves an $8\times$ theoretical speedup over the AR baseline while incurring only 3--7\% quality degradation across all metrics.
By contrast, T3D delivers only a $2\times$ theoretical speedup yet still incurs a 3--11\% quality drop, and dParallel achieves a $4.4\times$ speedup at the cost of 18--33\% degradation in clinical fidelity metrics.
\oursblkfour further provides a more conservative working point, where a $4\times$ theoretical speedup reduces clinical fidelity degradation to within 3\% on average.
Taken together, these results confirm that DCD achieves a consistently superior quality-speed trade-off across both block size configurations and against all existing distillation methods.

\begin{table}[t]
  \centering
  \caption{\textbf{Ablation on DCD components.} SW: step-wise token reweighting; CE: cross-entropy loss on the \texttt{<eos>} token; RKL: reverse KL divergence. Results are reported on CheXpert-Plus, ReXGradient, and MIMIC-CXR.}
  \label{tab:ablation_patch_feat}
  \setlength{\tabcolsep}{3.2pt}
  \small
  \resizebox{\linewidth}{!}{%
  \begin{tabular}{ccc cccc cccc cccc c}
    \toprule
    \multicolumn{3}{c}{\textbf{DCD settings}} & \multicolumn{4}{c}{\textbf{CheXpert-Plus}} & \multicolumn{4}{c}{\textbf{ReXGradient}} & \multicolumn{4}{c}{\textbf{MIMIC-CXR}} & \multirow{2}{*}{\textbf{PPL}} \\
    \cmidrule(lr){1-3} \cmidrule(lr){4-7} \cmidrule(lr){8-11} \cmidrule(lr){12-15}
    \textbf{SW} & \textbf{CE} & \textbf{RKL} & ROUGE-L & CIDEr & RateScore & SemScore & ROUGE-L & CIDEr & RateScore & SemScore & ROUGE-L & CIDEr & RateScore & SemScore \\
    \midrule
    &  &  & 52.57 & 3.61 & 54.87 & 46.93 & 63.28 & 5.17 & 61.32 & 58.36 & 51.73 & 3.63 & 51.19 & 43.50 & 23.72 \\
    $\checkmark$& & & 52.44 & 3.53 & 56.30 & 46.66 & 62.90 & 5.07 & 61.33 & 59.90 & 51.76 & 3.65 & 53.14 & 43.70 & 21.07 \\
    $\checkmark$ & $\checkmark$ &  & \textbf{56.14} & \textbf{4.20} & \textbf{57.40} & \textbf{49.57} & \textbf{66.39} & \textbf{5.54} & \textbf{62.18} & \textbf{66.28} & \textbf{54.12} & \textbf{4.05} & \textbf{53.95} & \textbf{45.57} & \textbf{18.83} \\
    $\checkmark$ & & $\checkmark$  & 52.51 & 3.41 & 55.20 & 45.32 & 61.71 & 4.94 & 61.26 & 59.55 & 50.91 & 3.48 & 53.31 & 42.43 & 20.23 \\
    $\checkmark$ & $\checkmark$ & $\checkmark$ & 53.25 & 3.72 & 57.25 & 47.07 & 63.82 & 5.21 & 61.47 & 63.46 & 51.48 & 3.67 & 53.42 & 43.86 & 21.32 \\
    \bottomrule
  \end{tabular}
  }
\end{table}

\subsection{Ablation Study of Distillation Component}

To investigate the individual contributions of the components in our proposed Direct Conditional Distillation (DCD) strategy, we conduct extensive ablation studies. Specifically, we evaluate the impact of Step-wise Weighting (SW), Cross-Entropy loss for the \texttt{<eos>} token (CE), and Reverse KL divergence (RKL). The results across three benchmarks are summarized in Tab.~\ref{tab:ablation_patch_feat}.

\noindent\textbf{Effect of Step-wise Weighting (SW).}
As shown in row 2 of Tab.~\ref{tab:ablation_patch_feat}, SW consistently improves generation stability across all datasets.
Upweighting tokens that are unmasked later in the denoising trajectory, which tend to carry stronger inter-token dependencies, reduces PPL from 23.72 to 21.07.
RaTEScore also improves steadily, e.g., from 54.87 to 56.30 on CheXpert-Plus, indicating that the reweighting helps the model better capture the conditioning provided by earlier committed tokens.

\noindent\textbf{Effect of \texttt{<eos>} Cross-Entropy Loss (CE).}
Adding CE supervision on the \texttt{<eos>} token produces the largest single improvement across all settings.
As discussed in Sec.~\ref{sec:RAD}, one-step diffusion models tend to assign low and unstable confidence to \texttt{<eos>} positions, leading to degenerate repetition loops.
Explicitly supervising the \texttt{<eos>} token with a one-hot target directly addresses this failure mode, resulting in a substantial gain in ROUGE-L from 52.44 to 56.14 on CheXpert-Plus and CIDEr from 3.65 to 4.05 on MIMIC-CXR.
PPL drops further to 18.83, its lowest value across all configurations, confirming that reliable termination is essential for overall report quality.

\noindent\textbf{Effect of Reverse KL (RKL).}
We also replace the forward KL objective with reverse KL (RKL) to examine whether its mode-seeking behavior benefits distillation.
As shown in rows 4 and 5 of Tab.~\ref{tab:ablation_patch_feat}, adding RKL consistently degrades performance.
Compared to the SW-only baseline, CIDEr drops from 3.65 to 3.48 and SemScore from 43.70 to 42.43 on MIMIC-CXR.
A likely cause is that clinical reports require the model to cover all plausible findings rather than concentrate probability mass on a single mode, which is the tendency encouraged by reverse KL.
Forward KL, which preserves the full teacher distribution, is therefore better suited for this task.

\subsection{Data Scale Analysis for RAD}
\label{exp:rad_data_analysis}
\begin{figure}[t]
  \centering
  \includegraphics[width=\linewidth]{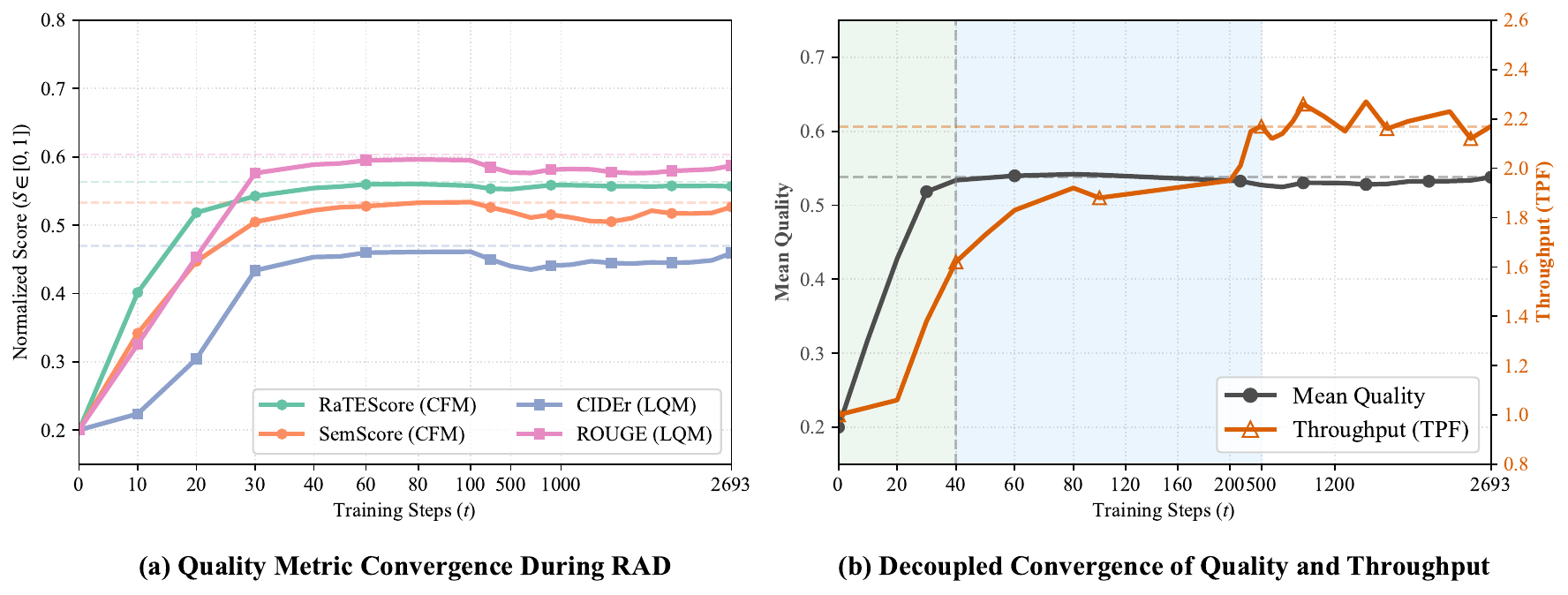}
  \caption{\textbf{Effect of training data scale during RAD adaptation.}
  \textbf{(a) Quality Metric Convergence During RAD.} Each colored curve tracks one of four clinical and linguistic metrics across training steps, and dashed horizontal lines mark the corresponding \ar baseline values. 
  \textbf{(b) Decoupled Convergence of Quality and Throughput.} The black curve shows the mean quality score and the red curve shows inference throughput (TPF), and dashed horizontal lines mark respective final values.}
  \label{fig:data-scale-diffusion-adaptation}
\end{figure}

We study how the amount of training data used in the RAD stage affects both model quality and inference throughput, by evaluating checkpoints at different training steps across all benchmarks.
Fig.~\ref{fig:data-scale-diffusion-adaptation} reports four clinical and linguistic metrics alongside inference throughput (TPF) as functions of training steps.

As shown in Fig.~\ref{fig:data-scale-diffusion-adaptation}(a), all four quality metrics follow a rapid climb-and-plateau pattern.
Within approximately 60 training steps, performance on all metrics reaches or surpasses the \ar baseline, indicating that RAD transfers the domain knowledge of the pretrained AR model efficiently and requires only a small fraction of the full training data to do so.

Fig.~\ref{fig:data-scale-diffusion-adaptation}(b) reveals that quality and throughput converge at different rates.
Quality saturates early (around step 60, corresponding to only 2.2\% of the full RAD training data), whereas TPF continues to improve throughout training, increasing from 1.62 to 2.17 ($+$33.95\%).
A plausible interpretation is that while cross-modal semantic alignment is established quickly, additional training is needed for the model to stabilize the joint token distribution within each block, which in turn enables more tokens to be committed with high confidence per denoising step and raises throughput.

\subsection{Effectiveness of DCD Against Native One-Step Decoding}
\label{sec:native_onestep}
To isolate the contribution of DCD, we compare \ours against \base under \textit{native one-step decoding}, where \base is forced to generate all response tokens simultaneously in a single forward pass without any distillation.
This setting directly exposes the mean-field bias inherent to block diffusion models and provides a controlled baseline for measuring the gain attributable to DCD alone.

\begin{table}[t]
  \centering
  \caption{\textbf{Comparison of \ours and \base under native one-step decoding.}
  Green percentages indicate the relative quality improvement of \ours over \base across three benchmarks.
  \textbf{Bold} values mark the higher score between the two model types within each block size and metric.}
  \label{tab:native_one-step_generation}
  \setlength{\tabcolsep}{3.5pt}
  \small
  \resizebox{\linewidth}{!}{%
\begin{tabular}{l l cccc cccc cccc}
    \toprule
    \multirow{2}{*}{\textbf{Block size}} & \multirow{2}{*}{\textbf{Model type}} & \multicolumn{4}{c}{\textbf{CheXpert-Plus}} & \multicolumn{4}{c}{\textbf{ReXGradient}} & \multicolumn{4}{c}{\textbf{MIMIC-CXR}} \\
    \cmidrule(lr){3-6} \cmidrule(lr){7-10} \cmidrule(lr){11-14}
    & & ROUGE-L & CIDEr & RateScore & SemScore & ROUGE-L & CIDEr & RateScore & SemScore & ROUGE-L & CIDEr & RateScore & SemScore \\
    \midrule
    \multirow{2}{*}{Block 4} & \base & 41.39 & 2.91 & 56.39 & 39.17 & 53.60 & 4.32 & 61.72 & 57.42 & 41.50 & 2.76 & 53.86 & 35.41 \\
    & \ours & \textbf{56.14}\smash{\rlap{$^{\textcolor{darkgreen}{36\%\uparrow}}$}} & \textbf{4.20}\smash{\rlap{$^{\textcolor{darkgreen}{44\%\uparrow}}$}} & \textbf{57.40}\smash{\rlap{$^{\textcolor{darkgreen}{2\%\uparrow}}$}} & \textbf{49.57}\smash{\rlap{$^{\textcolor{darkgreen}{27\%\uparrow}}$}} & \textbf{66.39}\smash{\rlap{$^{\textcolor{darkgreen}{24\%\uparrow}}$}} & \textbf{5.54}\smash{\rlap{$^{\textcolor{darkgreen}{28\%\uparrow}}$}} & \textbf{62.18}\smash{\rlap{$^{\textcolor{darkgreen}{1\%\uparrow}}$}} & \textbf{66.28}\smash{\rlap{$^{\textcolor{darkgreen}{15\%\uparrow}}$}} & \textbf{54.12}\smash{\rlap{$^{\textcolor{darkgreen}{30\%\uparrow}}$}} & \textbf{4.05}\smash{\rlap{$^{\textcolor{darkgreen}{47\%\uparrow}}$}} & \textbf{53.95}\smash{\rlap{$^{\textcolor{darkgreen}{0\%\uparrow}}$}} & \textbf{45.57}\smash{\rlap{$^{\textcolor{darkgreen}{29\%\uparrow}}$}} \\
    \midrule
    \multirow{2}{*}{Block 8} & \base & 41.27 & 2.81 & 52.12 & 38.74 & 53.23 & 4.19 & 59.92 & 54.29 & 41.08 & 2.63 & 51.34 & 34.18 \\
    & \ours  & \textbf{55.21}\smash{\rlap{$^{\textcolor{darkgreen}{34\%\uparrow}}$}} & \textbf{4.14}\smash{\rlap{$^{\textcolor{darkgreen}{47\%\uparrow}}$}} & \textbf{56.85}\smash{\rlap{$^{\textcolor{darkgreen}{9\%\uparrow}}$}} & \textbf{51.40}\smash{\rlap{$^{\textcolor{darkgreen}{33\%\uparrow}}$}} & \textbf{65.13}\smash{\rlap{$^{\textcolor{darkgreen}{22\%\uparrow}}$}} & \textbf{5.48}\smash{\rlap{$^{\textcolor{darkgreen}{31\%\uparrow}}$}} & \textbf{60.71}\smash{\rlap{$^{\textcolor{darkgreen}{1\%\uparrow}}$}} & \textbf{64.00}\smash{\rlap{$^{\textcolor{darkgreen}{18\%\uparrow}}$}} & \textbf{53.07}\smash{\rlap{$^{\textcolor{darkgreen}{29\%\uparrow}}$}} & \textbf{4.00}\smash{\rlap{$^{\textcolor{darkgreen}{52\%\uparrow}}$}} & \textbf{52.97}\smash{\rlap{$^{\textcolor{darkgreen}{3\%\uparrow}}$}} & \textbf{44.83}\smash{\rlap{$^{\textcolor{darkgreen}{31\%\uparrow}}$}} \\
    \bottomrule
\end{tabular}%
  }
\end{table}

As shown in Tab.~\ref{tab:native_one-step_generation}, DCD yields consistent and substantial improvements across both block sizes and all three benchmarks.
For \oursblkfour on CheXpert-Plus, ROUGE-L improves by 36\% and CIDEr by 44\% over the native one-step baseline.
SemScore increases by up to 29\% on MIMIC-CXR, indicating that DCD not only restores surface-level fluency but also substantially recovers the model's ability to identify positive pathological findings.
These gains are consistent across both block sizes, confirming that the benefit of DCD is not specific to a particular block configuration.

\subsection{Analysis on Normalization of Reports}
\label{exp:report_normalization_benefits}
To assess the impact of report normalization, we retrain the pipeline at each stage using unnormalized raw reports while keeping all other settings unchanged.
Tab.~\ref{tab:ablation_stage_datatype} summarizes the results across all three training stages.

\noindent\textbf{Impact on Stage I (Continued Pre-training).}
Normalized data yields substantial improvements in Stage I across all metrics.
On CheXpert-Plus, ROUGE-L increases from 23.82 to 56.89 and RaTEScore from 45.59 to 59.18.
This large gap indicates that raw reports, which frequently contain inconsistent phrasing and systematically omit negative findings, provide ambiguous supervision that prevents the model from learning reliable visual-textual mappings.

\noindent\textbf{Impact on Stage II and Stage III (Adaptation and Distillation).}
The gap persists through Stage II and widens further in Stage III.
Without normalization, ROUGE-L collapses to 18.79 on CheXpert-Plus and 19.98 on ReXGradient after distillation, substantially below the already-degraded Stage II unnormalized baseline.
With normalization, the model retains strong clinical accuracy across all stages, with SemScore remaining above 45.0 on CheXpert-Plus even after one-step distillation.
This progressive degradation under unnormalized training suggests that noisy supervision compounds across stages: the omission bias that distorts pretraining is further amplified by the mean-field bias of block diffusion, and becomes most severe when the student must reproduce the teacher's full conditional distribution in a single step.

\begin{table}[t]
  \centering
  \caption{\textbf{Ablation on report data types across all training stages.} Each stage contains two settings: normalized and unnormalized reports.}
  \label{tab:ablation_stage_datatype}
  \setlength{\tabcolsep}{3.2pt}
  \small
  \resizebox{\linewidth}{!}{%
  \begin{tabular}{ll cccc cccc cccc}
    \toprule
    \multirow{2}{*}{\textbf{Stage}} & \multirow{2}{*}{\textbf{Data Type}} & \multicolumn{4}{c}{\textbf{CheXpert-Plus}} & \multicolumn{4}{c}{\textbf{ReXGradient}} & \multicolumn{4}{c}{\textbf{MIMIC-CXR}} \\
    \cmidrule(lr){3-6} \cmidrule(lr){7-10} \cmidrule(lr){11-14}
    & & ROUGE-L & CIDEr & RateScore & SemScore & ROUGE-L & CIDEr & RateScore & SemScore & ROUGE-L & CIDEr & RateScore & SemScore \\
    \midrule
    \multirow{2}{*}{\textbf{Stage I}} & Normalized & \textbf{56.89} & \textbf{4.47} & \textbf{59.18} & \textbf{52.94} & \textbf{67.07} & \textbf{5.68} & \textbf{64.39} & \textbf{66.60} & \textbf{54.55} & \textbf{4.32} & \textbf{56.58} & \textbf{46.70} \\
    & Unnormalized & 23.82 & 1.53 & 45.59 & 32.44 & 25.67 & 1.49 & 26.05 & 40.77 & 25.06 & 1.56 & 42.72 & 33.54\\
    \midrule
    \multirow{2}{*}{\textbf{Stage II}} & Normalized & \textbf{56.90} & \textbf{4.25} & \textbf{59.12} & \textbf{52.64} & \textbf{66.46} & \textbf{5.64} & \textbf{63.02} & \textbf{66.68} & \textbf{54.45} & \textbf{4.16} & \textbf{55.85} & \textbf{46.10} \\
    & Unnormalized & 23.47 & 1.51 & 39.60 & 30.46 & 24.97 & 1.45 & 23.90 & 36.97 & 24.54 & 1.53 & 37.76 & 32.64 \\
    \midrule
    \multirow{2}{*}{\textbf{Stage III}} & Normalized & \textbf{56.14} & \textbf{4.20} & \textbf{57.40} & \textbf{49.57} & \textbf{66.39} & \textbf{5.54} & \textbf{62.18} & \textbf{66.28} & \textbf{54.12} & \textbf{4.05} & \textbf{53.95} & \textbf{45.57} \\
    & Unnormalized & 18.79 & 1.19 & 42.08 & 27.53 & 19.98 & 1.20 & 24.98 & 31.70 & 19.61 & 1.24 & 37.55 & 28.02 \\
    \bottomrule
  \end{tabular}%
  }
\end{table}

\enlargethispage{2\baselineskip}
\section{Conclusion}
\label{sec:conclusion}

We presented \ours, a discrete diffusion VLM that demonstrates high clinical accuracy and extreme decoding efficiency are simultaneously achievable in automated chest X-ray reporting.
By rethinking both the AR-to-diffusion conversion and the distillation target, \ours reduces training cost, eliminates multi-step denoising, and consistently outperforms autoregressive and diffusion-based state-of-the-art methods across standard benchmarks.
We hope this work provides a strong foundation for efficient and reliable generation in broader medical multimodal settings beyond CXR reporting.

\clearpage
\bibliographystyle{plainnat}
\bibliography{main}

\clearpage

\beginappendix

\setcounter{figure}{0}
\setcounter{table}{0}
\renewcommand{\thefigure}{A.\arabic{figure}}
\renewcommand{\thetable}{A.\arabic{table}}

This supplementary material provides additional details and results to complement the main paper, organized as follows.
\begin{itemize}
    \item \textbf{Sec.~\ref{sec:appendix-examples}: Qualitative Examples.} We present qualitative examples illustrating the effect of DCD on generation quality and the diagnostic capability of \ours.
    \item \textbf{Sec.~\ref{sec:appendix-experiment}: Experimental Details.} We describe the data preprocessing pipeline (Sec.~\ref{sec:appendix-data-details}), training configurations for all three stages (Sec.~\ref{sec:appendix-training-details}), definitions and adaptations of all evaluation metrics (Sec.~\ref{sec:metric_details}), and the prompt templates used throughout the study (Sec.~\ref{sec:appendix-prompts}).
    \item \textbf{Sec.~\ref{sec:appendix-baselines}: Implementation of Baselines.} We provide implementation details of all baseline methods, including hyperparameter settings and training loss combinations.
    \item \textbf{Sec.~\ref{sec:appendix-detailed-results}: Detailed Results.} We report extended quantitative results with per-dataset breakdowns across CheXpert-Plus, ReXGradient, and MIMIC-CXR.
    \item \textbf{Sec.~\ref{sec:proof_fused_kv}: Fused Block KV Cache Analysis.} We formally show that Fused Block KV Cache preserves total FLOPs while halving the number of forward passes for one-step-per-block decoding.
\end{itemize}

\section{Qualitative examples}
\label{sec:appendix-examples}
\begin{figure*}[p]
    \centering
    \includegraphics[width=\linewidth, height=0.95\textheight, keepaspectratio]{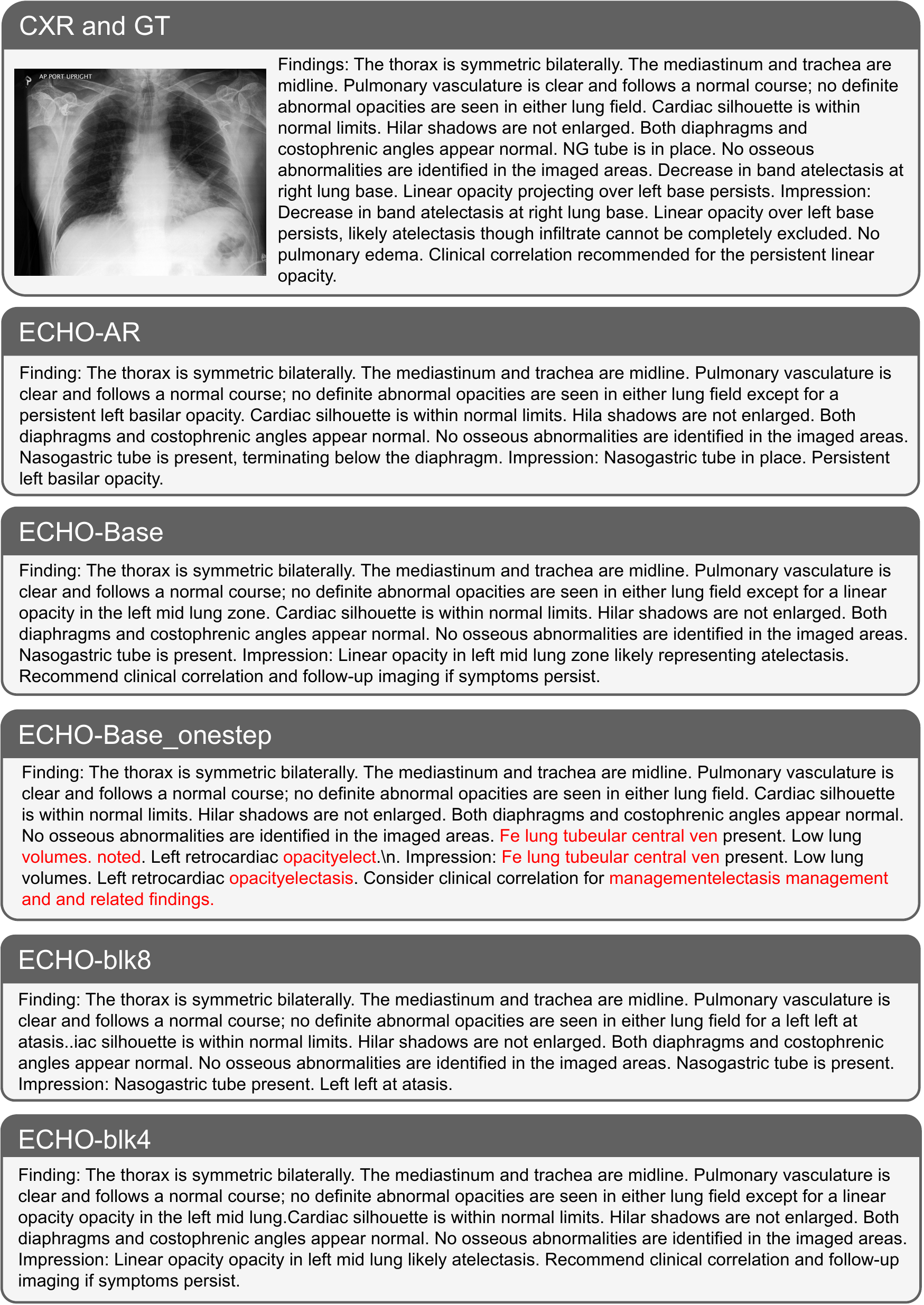}
    \caption{Qualitative comparison of report generation quality across model variants.
    \texttt{ECHO-Base\_{onestep}} denotes \base forced to decode in a single step without distillation.}
    \label{fig:quality1}
\end{figure*}

\begin{figure*}[p]
    \centering
    \includegraphics[width=\linewidth, height=0.95\textheight, keepaspectratio]{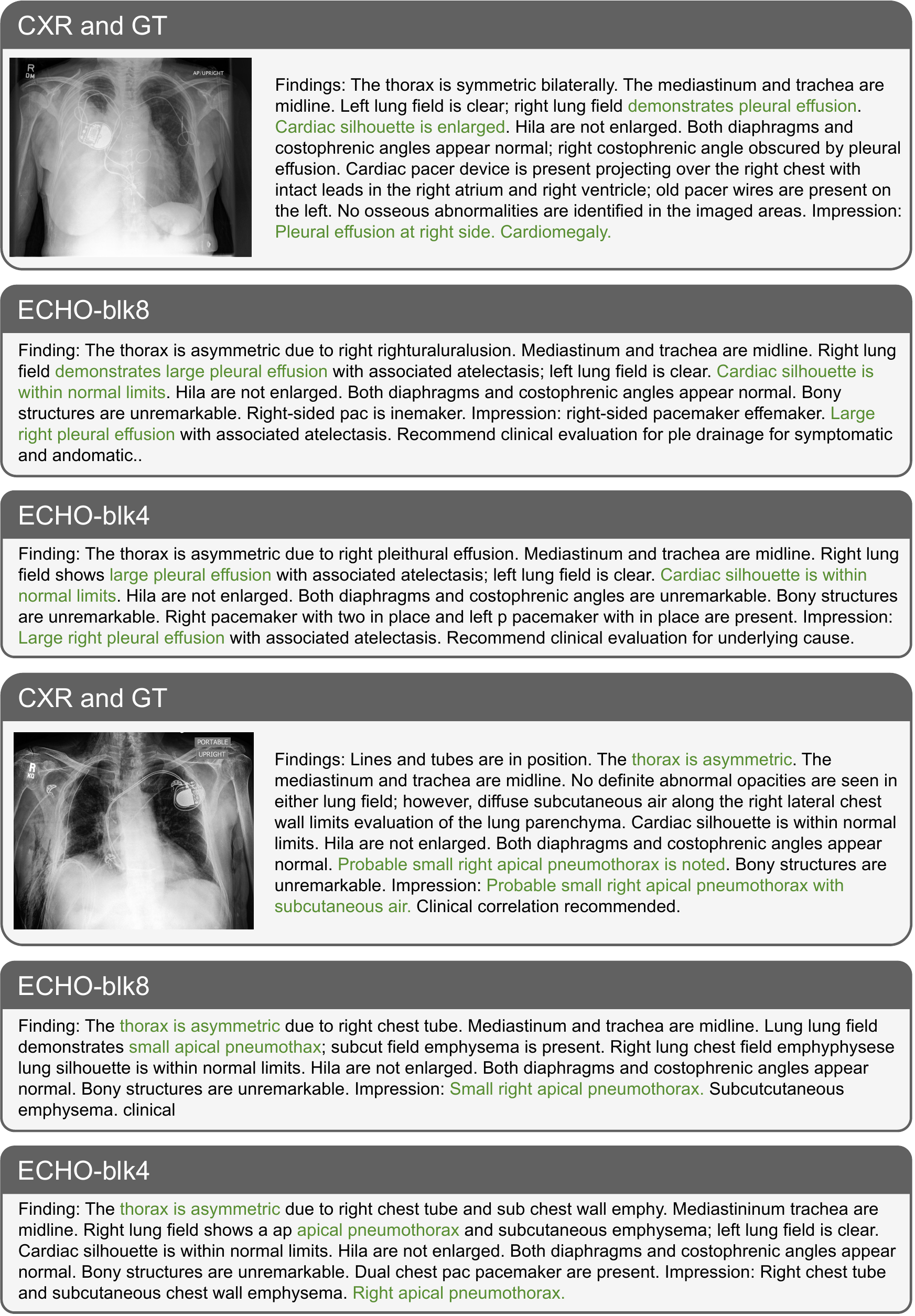}
    \caption{Qualitative examples of positive pathology detection by \ours.
    Correctly identified abnormal findings are highlighted in green.}
    \label{fig:quality2}
\end{figure*}

We present two sets of qualitative comparisons.
Fig.~\ref{fig:quality1} analyzes differences in generation quality across model variants, and Fig.~\ref{fig:quality2} demonstrates the ability of \ours to detect positive pathological findings.

Fig.~\ref{fig:quality1} compares reports generated by \texttt{ECHO-AR}, \base, \texttt{ECHO-Base\_{onestep}}, \oursblkeight, and \oursblkfour, where \texttt{ECHO-Base\_{onestep}} denotes \base forced to decode in a single step without distillation.
It is obvious that \base produces fluent and coherent reports without noticeable errors.
In contrast, \texttt{ECHO-Base\_{onestep}} exhibits severe token disorder and repetition caused by mean-field factorization.
After applying DCD, \oursblkeight and \oursblkfour largely eliminate these artifacts, with only minor residual repetition.
This comparison confirms that DCD effectively mitigates mean-field bias and yields more stable one-step generation.

Fig.~\ref{fig:quality2} shows examples where \ours correctly identifies abnormal findings, with correct predictions highlighted in green.
Across these cases, \ours accurately detects multiple abnormalities and describes them using appropriate clinical terminology, demonstrating reliable diagnostic capability for CXR report generation.


\begin{table*}[t]
  \centering
  \caption{Detailed per-metric comparison on CheXpert-Plus across Chinese and English evaluation.
  Green percentages indicate the relative performance drop of \ours compared to \ar.
  Bold values denote the best result among all distillation methods.}
  \label{tab:performance_language_split_chexpertplus}
  \setlength{\tabcolsep}{3.5pt}
  \small
  \resizebox{\linewidth}{!}{%
\begin{tabular}{l ccccc ccccc}
    \toprule
    \multirow{2}{*}{\textbf{Methods}} & \multicolumn{5}{c}{\textbf{Chinese}} & \multicolumn{5}{c}{\textbf{English}} \\
    \cmidrule(lr){2-6} \cmidrule(lr){7-11}
    & ROUGE-L & METEOR & CIDEr & RateScore & SemScore & ROUGE-L & METEOR & CIDEr & RateScore & SemScore \\
    \midrule
    \multicolumn{11}{c}{\textbf{Proprietary general models}} \\
    \midrule
    Gemini3-Pro~\cite{deepmind_gemini3pro_modelcard_2025} & 26.79 & 29.40 & 2.14 & 41.87 & 32.21 & 27.12 & 25.71 & 0.92 & 39.84 & 27.23 \\
    Qwen3-Max~\cite{bai2025qwen3} & 26.33 & 30.17 & 2.12 & 43.20 & 29.88 & 28.04 & 26.82 & 0.93 & 40.06 & 28.60 \\
    \midrule
    \multicolumn{11}{c}{\textbf{Autoregressive Medical Models}} \\
    \midrule
    LLaVA-Med~\cite{li2023llava} & 10.58 & 5.50 & 1.13 & 10.03 & 11.69 & 3.25 & 0.56 & 0.18 & 10.00 & 11.09 \\
    Lingshu-7B~\cite{xu2025lingshu} & 19.96 & 16.81 & 1.50 & 28.46 & 26.51 & 23.93 & 14.19 & 0.73 & 34.22 & 28.56 \\
    Hulu-Med-7B~\cite{jiang2025hulu} & 21.18 & 18.45 & 1.67 & 22.09 & 27.34 & 23.57 & 14.34 & 0.63 & 30.41 & 19.52 \\
    MedGemma-27B~\cite{sellergren2025medgemma} & 15.56 & 20.76 & 0.95 & 34.83 & 29.91 & 25.04 & 23.59 & 0.62 & 42.84 & 33.68 \\
    Lingshu-32B~\cite{xu2025lingshu} & 16.61 & 12.92 & 1.39 & 27.68 & 21.25 & 24.19 & 15.86 & 0.73 & 34.16 & 27.39 \\
    Hulu-Med-32B~\cite{jiang2025hulu} & 19.93 & 14.37 & 1.45 & 22.85 & 22.57 & 19.00 & 11.06 & 0.47 & 31.49 & 21.57 \\
    \midrule
    \multicolumn{11}{c}{\textbf{Diffusion Medical Methods}} \\
    \midrule
    LLaDA-MedV~\cite{dong2025llada} & 2.79 & 3.13 & 0.16 & 34.83 & 18.21 & 12.59 & 11.55 & 0.28 & 18.60 & 17.31 \\
    \base & 55.30 & 54.52 & 4.62 & 54.68 & 50.69 & 58.50 & 52.02 & 3.88 & 63.57 & 54.59 \\
    CD4LM~\cite{liang2026cd4lm}   & 44.33\smash{\rlap{$^{\textcolor{red}{20\%\downarrow}}$}} & 49.57\smash{\rlap{$^{\textcolor{red}{9\%\downarrow}}$}} & 3.70\smash{\rlap{$^{\textcolor{red}{20\%\downarrow}}$}} & 51.87\smash{\rlap{$^{\textcolor{red}{5\%\downarrow}}$}} & 44.74\smash{\rlap{$^{\textcolor{red}{12\%\downarrow}}$}} & 54.09\smash{\rlap{$^{\textcolor{red}{8\%\downarrow}}$}} & 47.25\smash{\rlap{$^{\textcolor{red}{9\%\downarrow}}$}} & 3.07\smash{\rlap{$^{\textcolor{red}{21\%\downarrow}}$}} & 59.94\smash{\rlap{$^{\textcolor{red}{6\%\downarrow}}$}} & 42.98\smash{\rlap{$^{\textcolor{red}{21\%\downarrow}}$}} \\
    d3LLM~\cite{qian2026d3llm}   & 34.59\smash{\rlap{$^{\textcolor{red}{37\%\downarrow}}$}} & 45.14\smash{\rlap{$^{\textcolor{red}{17\%\downarrow}}$}} & 2.77\smash{\rlap{$^{\textcolor{red}{40\%\downarrow}}$}} & 51.20\smash{\rlap{$^{\textcolor{red}{6\%\downarrow}}$}} & 39.13\smash{\rlap{$^{\textcolor{red}{23\%\downarrow}}$}} & 53.49\smash{\rlap{$^{\textcolor{red}{9\%\downarrow}}$}} & 47.45\smash{\rlap{$^{\textcolor{red}{9\%\downarrow}}$}} & 3.08\smash{\rlap{$^{\textcolor{red}{21\%\downarrow}}$}} & 59.57\smash{\rlap{$^{\textcolor{red}{6\%\downarrow}}$}} & 41.57\smash{\rlap{$^{\textcolor{red}{24\%\downarrow}}$}} \\
    dParallel~\cite{chen2025dparallel}   & 40.61\smash{\rlap{$^{\textcolor{red}{27\%\downarrow}}$}} & 41.94\smash{\rlap{$^{\textcolor{red}{23\%\downarrow}}$}} & 3.47\smash{\rlap{$^{\textcolor{red}{25\%\downarrow}}$}} & 36.03\smash{\rlap{$^{\textcolor{red}{34\%\downarrow}}$}} & 35.70\smash{\rlap{$^{\textcolor{red}{30\%\downarrow}}$}} & 43.83\smash{\rlap{$^{\textcolor{red}{25\%\downarrow}}$}} & 38.94\smash{\rlap{$^{\textcolor{red}{25\%\downarrow}}$}} & 2.87\smash{\rlap{$^{\textcolor{red}{26\%\downarrow}}$}} & 44.91\smash{\rlap{$^{\textcolor{red}{29\%\downarrow}}$}} & 37.74\smash{\rlap{$^{\textcolor{red}{31\%\downarrow}}$}} \\
    T3D~\cite{zhang2026t3d}   & 51.71\smash{\rlap{$^{\textcolor{red}{6\%\downarrow}}$}} & 54.69\smash{\rlap{$^{\textcolor{red}{0\%\downarrow}}$}} & 4.38\smash{\rlap{$^{\textcolor{red}{5\%\downarrow}}$}} & 52.68\smash{\rlap{$^{\textcolor{red}{4\%\downarrow}}$}} & 48.11\smash{\rlap{$^{\textcolor{red}{5\%\downarrow}}$}} & 58.29\smash{\rlap{$^{\textcolor{red}{0\%\downarrow}}$}} & 51.84\smash{\rlap{$^{\textcolor{red}{0\%\downarrow}}$}} & 3.75\smash{\rlap{$^{\textcolor{red}{3\%\downarrow}}$}} & 61.12\smash{\rlap{$^{\textcolor{red}{4\%\downarrow}}$}} & 51.63\smash{\rlap{$^{\textcolor{red}{5\%\downarrow}}$}} \\
    \oursblkeight   & 52.45\smash{\rlap{$^{\textcolor{darkgreen}{5\%\downarrow}}$}} & 51.67\smash{\rlap{$^{\textcolor{darkgreen}{5\%\downarrow}}$}} & 4.47\smash{\rlap{$^{\textcolor{darkgreen}{3\%\downarrow}}$}} & 52.95\smash{\rlap{$^{\textcolor{darkgreen}{3\%\downarrow}}$}} & \textbf{48.89}\smash{\rlap{$^{\textcolor{darkgreen}{4\%\downarrow}}$}} & 57.98\smash{\rlap{$^{\textcolor{darkgreen}{1\%\downarrow}}$}} & 49.20\smash{\rlap{$^{\textcolor{darkgreen}{5\%\downarrow}}$}} & 3.80\smash{\rlap{$^{\textcolor{darkgreen}{2\%\downarrow}}$}} & 60.75\smash{\rlap{$^{\textcolor{darkgreen}{4\%\downarrow}}$}} & \textbf{53.92}\smash{\rlap{$^{\textcolor{darkgreen}{1\%\downarrow}}$}} \\
    \oursblkfour   & \textbf{54.66}\smash{\rlap{$^{\textcolor{darkgreen}{1\%\downarrow}}$}} & \textbf{53.15}\smash{\rlap{$^{\textcolor{darkgreen}{3\%\downarrow}}$}} & \textbf{4.59}\smash{\rlap{$^{\textcolor{darkgreen}{1\%\downarrow}}$}} & \textbf{53.76}\smash{\rlap{$^{\textcolor{darkgreen}{2\%\downarrow}}$}} & 48.11\smash{\rlap{$^{\textcolor{darkgreen}{5\%\downarrow}}$}} & \textbf{57.62}\smash{\rlap{$^{\textcolor{darkgreen}{2\%\downarrow}}$}} & \textbf{50.46}\smash{\rlap{$^{\textcolor{darkgreen}{3\%\downarrow}}$}} & \textbf{3.81}\smash{\rlap{$^{\textcolor{darkgreen}{2\%\downarrow}}$}} & \textbf{61.04}\smash{\rlap{$^{\textcolor{darkgreen}{4\%\downarrow}}$}} & 51.03\smash{\rlap{$^{\textcolor{darkgreen}{7\%\downarrow}}$}} \\
    \bottomrule
\end{tabular}%
  }
\end{table*}


\begin{table*}[t]
  \centering
  \caption{Detailed per-metric comparison on RexGradient across Chinese and English evaluation.
  Green percentages indicate the relative performance drop of \ours compared to \ar.
  Bold values denote the best result among all distillation methods.}
  \label{tab:performance_language_split_rexgradient}
  \setlength{\tabcolsep}{3.5pt}
  \small
  \resizebox{\linewidth}{!}{%
\begin{tabular}{l ccccc ccccc}
    \toprule
    \multirow{2}{*}{\textbf{Methods}} & \multicolumn{5}{c}{\textbf{Chinese}} & \multicolumn{5}{c}{\textbf{English}} \\
    \cmidrule(lr){2-6} \cmidrule(lr){7-11}
    & ROUGE-L & METEOR & CIDEr & RateScore & SemScore & ROUGE-L & METEOR & CIDEr & RateScore & SemScore \\
    \midrule
    \multicolumn{11}{c}{\textbf{Proprietary general models}} \\
    \midrule
    Gemini3-Pro~\cite{deepmind_gemini3pro_modelcard_2025} & 34.10 & 36.70 & 2.60 & 41.78 & 42.59 & 30.10 & 26.34 & 1.03 & 25.26 & 35.46 \\
    Qwen3-Max~\cite{bai2025qwen3} & 31.32 & 36.62 & 2.48 & 45.86 & 39.13 & 29.67 & 29.84 & 0.97 & 32.38 & 37.14 \\
    \midrule
    \multicolumn{11}{c}{\textbf{Autoregressive Medical Models}} \\
    \midrule
    LLaVA-Med~\cite{li2023llava} & 11.35 & 5.92 & 1.23 & 10.01 & 18.46 & 3.37 & 0.62 & 0.21 & 10.00 & 15.62 \\
    Lingshu-7B~\cite{xu2025lingshu} & 22.24 & 19.27 & 1.68 & 20.99 & 32.56 & 25.01 & 14.87 & 0.83 & 19.51 & 37.07 \\
    Hulu-Med-7B~\cite{jiang2025hulu} & 21.35 & 18.96 & 1.70 & 22.48 & 24.36 & 21.96 & 14.48 & 0.58 & 23.06 & 22.76 \\
    MedGemma-27B~\cite{sellergren2025medgemma} & 16.97 & 23.29 & 1.08 & 24.16 & 34.87 & 29.85 & 26.64 & 0.72 & 27.84 & 40.61 \\
    Lingshu-32B~\cite{xu2025lingshu} & 18.45 & 14.29 & 1.53 & 21.07 & 27.75 & 25.75 & 20.06 & 0.87 & 19.58 & 35.22 \\
    Hulu-Med-32B~\cite{jiang2025hulu} & 21.46 & 17.15 & 1.63 & 22.72 & 28.06 & 23.17 & 15.49 & 0.59 & 23.30 & 25.28 \\
    \midrule
    \multicolumn{11}{c}{\textbf{Diffusion Medical Methods}} \\
    \midrule
    LLaDA-MedV~\cite{dong2025llada} & 2.87 & 3.91 & 0.13 & 28.05 & 14.46 & 16.57 & 14.30 & 0.34 & 8.33 & 29.73 \\
    \base & 64.59 & 64.53 & 5.65 & 53.09 & 61.64 & 68.32 & 65.18 & 5.63 & 72.95 & 71.72 \\
    CD4LM~\cite{liang2026cd4lm}   & 54.83\smash{\rlap{$^{\textcolor{red}{15\%\downarrow}}$}} & 58.33\smash{\rlap{$^{\textcolor{red}{10\%\downarrow}}$}} & 4.74\smash{\rlap{$^{\textcolor{red}{16\%\downarrow}}$}} & 48.56\smash{\rlap{$^{\textcolor{red}{9\%\downarrow}}$}} & 59.24\smash{\rlap{$^{\textcolor{red}{4\%\downarrow}}$}} & 62.62\smash{\rlap{$^{\textcolor{red}{8\%\downarrow}}$}} & 58.85\smash{\rlap{$^{\textcolor{red}{10\%\downarrow}}$}} & 4.14\smash{\rlap{$^{\textcolor{red}{26\%\downarrow}}$}} & 68.15\smash{\rlap{$^{\textcolor{red}{7\%\downarrow}}$}} & 52.00\smash{\rlap{$^{\textcolor{red}{27\%\downarrow}}$}} \\
    d3LLM~\cite{qian2026d3llm}   & 45.53\smash{\rlap{$^{\textcolor{red}{30\%\downarrow}}$}} & 53.42\smash{\rlap{$^{\textcolor{red}{17\%\downarrow}}$}} & 3.79\smash{\rlap{$^{\textcolor{red}{33\%\downarrow}}$}} & 46.53\smash{\rlap{$^{\textcolor{red}{12\%\downarrow}}$}} & 53.11\smash{\rlap{$^{\textcolor{red}{14\%\downarrow}}$}} & 62.82\smash{\rlap{$^{\textcolor{red}{8\%\downarrow}}$}} & 60.61\smash{\rlap{$^{\textcolor{red}{7\%\downarrow}}$}} & 4.56\smash{\rlap{$^{\textcolor{red}{19\%\downarrow}}$}} & 69.47\smash{\rlap{$^{\textcolor{red}{5\%\downarrow}}$}} & 53.88\smash{\rlap{$^{\textcolor{red}{25\%\downarrow}}$}} \\
    dParallel~\cite{chen2025dparallel}   & 47.26\smash{\rlap{$^{\textcolor{red}{27\%\downarrow}}$}} & 47.78\smash{\rlap{$^{\textcolor{red}{26\%\downarrow}}$}} & 4.09\smash{\rlap{$^{\textcolor{red}{28\%\downarrow}}$}} & 37.48\smash{\rlap{$^{\textcolor{red}{29\%\downarrow}}$}} & 45.91\smash{\rlap{$^{\textcolor{red}{26\%\downarrow}}$}} & 50.21\smash{\rlap{$^{\textcolor{red}{27\%\downarrow}}$}} & 48.29\smash{\rlap{$^{\textcolor{red}{26\%\downarrow}}$}} & 3.94\smash{\rlap{$^{\textcolor{red}{30\%\downarrow}}$}} & 53.52\smash{\rlap{$^{\textcolor{red}{27\%\downarrow}}$}} & 50.11\smash{\rlap{$^{\textcolor{red}{30\%\downarrow}}$}} \\
    T3D~\cite{zhang2026t3d}   & 61.56\smash{\rlap{$^{\textcolor{red}{5\%\downarrow}}$}} & 63.40\smash{\rlap{$^{\textcolor{red}{2\%\downarrow}}$}} & 5.32\smash{\rlap{$^{\textcolor{red}{6\%\downarrow}}$}} & 48.28\smash{\rlap{$^{\textcolor{red}{9\%\downarrow}}$}} & 59.25\smash{\rlap{$^{\textcolor{red}{4\%\downarrow}}$}} & 67.17\smash{\rlap{$^{\textcolor{red}{2\%\downarrow}}$}} & 64.58\smash{\rlap{$^{\textcolor{red}{1\%\downarrow}}$}} & 5.13\smash{\rlap{$^{\textcolor{red}{9\%\downarrow}}$}} & 68.69\smash{\rlap{$^{\textcolor{red}{6\%\downarrow}}$}} & 64.06\smash{\rlap{$^{\textcolor{red}{11\%\downarrow}}$}} \\
    \oursblkeight   & 62.44\smash{\rlap{$^{\textcolor{darkgreen}{3\%\downarrow}}$}} & 62.02\smash{\rlap{$^{\textcolor{darkgreen}{4\%\downarrow}}$}} & 5.44\smash{\rlap{$^{\textcolor{darkgreen}{4\%\downarrow}}$}} & 50.65\smash{\rlap{$^{\textcolor{darkgreen}{5\%\downarrow}}$}} & 60.36\smash{\rlap{$^{\textcolor{darkgreen}{2\%\downarrow}}$}} & 67.82\smash{\rlap{$^{\textcolor{darkgreen}{1\%\downarrow}}$}} & 64.05\smash{\rlap{$^{\textcolor{darkgreen}{2\%\downarrow}}$}} & 5.53\smash{\rlap{$^{\textcolor{darkgreen}{2\%\downarrow}}$}} & 69.20\smash{\rlap{$^{\textcolor{darkgreen}{5\%\downarrow}}$}} & 67.64\smash{\rlap{$^{\textcolor{darkgreen}{6\%\downarrow}}$}} \\
    \oursblkfour   & \textbf{64.74}\smash{\rlap{$^{\textcolor{darkgreen}{0\%\downarrow}}$}} & \textbf{63.17}\smash{\rlap{$^{\textcolor{darkgreen}{2\%\downarrow}}$}} & \textbf{5.51}\smash{\rlap{$^{\textcolor{darkgreen}{2\%\downarrow}}$}} & \textbf{52.69}\smash{\rlap{$^{\textcolor{darkgreen}{1\%\downarrow}}$}} & \textbf{61.60}\smash{\rlap{$^{\textcolor{darkgreen}{0\%\downarrow}}$}} & \textbf{68.04}\smash{\rlap{$^{\textcolor{darkgreen}{0\%\downarrow}}$}} & \textbf{64.65}\smash{\rlap{$^{\textcolor{darkgreen}{1\%\downarrow}}$}} & \textbf{5.57}\smash{\rlap{$^{\textcolor{darkgreen}{1\%\downarrow}}$}} & \textbf{71.67}\smash{\rlap{$^{\textcolor{darkgreen}{2\%\downarrow}}$}} & \textbf{70.96}\smash{\rlap{$^{\textcolor{darkgreen}{1\%\downarrow}}$}} \\
    \bottomrule
\end{tabular}%
  }
\end{table*}


\begin{table*}[t]
  \centering
  \caption{Detailed per-metric comparison on MIMIC across Chinese and English evaluation.
  Green percentages indicate the relative performance drop of \ours compared to \ar.
  Bold values denote the best result among all distillation methods.}
  \label{tab:performance_language_split_mimic}
  \setlength{\tabcolsep}{3.5pt}
  \small
  \resizebox{\linewidth}{!}{%
\begin{tabular}{l ccccc ccccc}
    \toprule
    \multirow{2}{*}{\textbf{Methods}} & \multicolumn{5}{c}{\textbf{Chinese}} & \multicolumn{5}{c}{\textbf{English}} \\
    \cmidrule(lr){2-6} \cmidrule(lr){7-11}
    & ROUGE-L & METEOR & CIDEr & RateScore & SemScore & ROUGE-L & METEOR & CIDEr & RateScore & SemScore \\
    \midrule
    \multicolumn{11}{c}{\textbf{Proprietary general models}} \\
    \midrule
    Gemini3-Pro~\cite{deepmind_gemini3pro_modelcard_2025} & 25.86 & 28.37 & 2.13 & 41.46 & 30.62 & 28.30 & 26.28 & 0.98 & 41.38 & 30.28 \\
    Qwen3-Max~\cite{bai2025qwen3} & 25.62 & 29.10 & 2.08 & 41.93 & 26.95 & 28.10 & 27.21 & 0.94 & 39.64 & 27.69 \\
    \midrule
    \multicolumn{11}{c}{\textbf{Autoregressive Medical Models}} \\
    \midrule
    LLaVA-Med~\cite{li2023llava} & 10.32 & 5.42 & 1.12 & 10.02 & 10.42 & 3.05 & 0.55 & 0.18 & 10.00 & 8.02 \\
    Lingshu-7B~\cite{xu2025lingshu} & 18.93 & 16.27 & 1.48 & 28.69 & 24.04 & 25.55 & 14.79 & 0.83 & 36.69 & 29.84 \\
    Hulu-Med-7B~\cite{jiang2025hulu} & 21.64 & 16.37 & 1.68 & 24.36 & 23.56 & 22.88 & 14.78 & 0.68 & 21.41 & 21.04 \\
    MedGemma-27B~\cite{sellergren2025medgemma} & 15.46 & 20.82 & 0.98 & 33.74 & 27.35 & 26.14 & 24.06 & 0.65 & 43.22 & 33.13 \\
    Lingshu-32B~\cite{xu2025lingshu} & 16.52 & 12.63 & 1.41 & 28.35 & 19.65 & 24.94 & 14.61 & 0.79 & 36.45 & 28.62 \\
    Hulu-Med-32B~\cite{jiang2025hulu} & 18.25 & 13.51 & 1.46 & 26.36 & 15.41 & 17.66 & 10.53 & 0.52 & 22.72 & 21.06 \\
    \midrule
    \multicolumn{11}{c}{\textbf{Diffusion Medical Methods}} \\
    \midrule
    LLaDA-MedV~\cite{dong2025llada} & 2.86 & 3.85 & 0.13 & 37.70 & 15.95 & 16.33 & 13.18 & 0.33 & 17.24 & 18.57 \\
    \base & 52.59 & 52.08 & 4.51 & 50.98 & 43.63 & 56.32 & 51.40 & 3.81 & 60.71 & 48.58 \\
    CD4LM~\cite{liang2026cd4lm}   & 40.85\smash{\rlap{$^{\textcolor{red}{22\%\downarrow}}$}} & 46.58\smash{\rlap{$^{\textcolor{red}{11\%\downarrow}}$}} & 3.43\smash{\rlap{$^{\textcolor{red}{24\%\downarrow}}$}} & 47.16\smash{\rlap{$^{\textcolor{red}{7\%\downarrow}}$}} & 39.78\smash{\rlap{$^{\textcolor{red}{9\%\downarrow}}$}} & 51.56\smash{\rlap{$^{\textcolor{red}{8\%\downarrow}}$}} & 45.33\smash{\rlap{$^{\textcolor{red}{12\%\downarrow}}$}} & 2.83\smash{\rlap{$^{\textcolor{red}{26\%\downarrow}}$}} & 57.07\smash{\rlap{$^{\textcolor{red}{6\%\downarrow}}$}} & 37.74\smash{\rlap{$^{\textcolor{red}{22\%\downarrow}}$}} \\
    d3LLM~\cite{qian2026d3llm}   & 30.73\smash{\rlap{$^{\textcolor{red}{42\%\downarrow}}$}} & 42.33\smash{\rlap{$^{\textcolor{red}{19\%\downarrow}}$}} & 2.47\smash{\rlap{$^{\textcolor{red}{45\%\downarrow}}$}} & 45.59\smash{\rlap{$^{\textcolor{red}{11\%\downarrow}}$}} & 30.14\smash{\rlap{$^{\textcolor{red}{31\%\downarrow}}$}} & 51.54\smash{\rlap{$^{\textcolor{red}{8\%\downarrow}}$}} & 45.60\smash{\rlap{$^{\textcolor{red}{11\%\downarrow}}$}} & 2.82\smash{\rlap{$^{\textcolor{red}{26\%\downarrow}}$}} & 56.41\smash{\rlap{$^{\textcolor{red}{7\%\downarrow}}$}} & 38.26\smash{\rlap{$^{\textcolor{red}{21\%\downarrow}}$}} \\
    dParallel~\cite{chen2025dparallel}   & 37.84\smash{\rlap{$^{\textcolor{red}{28\%\downarrow}}$}} & 39.49\smash{\rlap{$^{\textcolor{red}{24\%\downarrow}}$}} & 3.23\smash{\rlap{$^{\textcolor{red}{28\%\downarrow}}$}} & 36.03\smash{\rlap{$^{\textcolor{red}{29\%\downarrow}}$}} & 27.95\smash{\rlap{$^{\textcolor{red}{36\%\downarrow}}$}} & 42.12\smash{\rlap{$^{\textcolor{red}{25\%\downarrow}}$}} & 37.72\smash{\rlap{$^{\textcolor{red}{27\%\downarrow}}$}} & 2.69\smash{\rlap{$^{\textcolor{red}{29\%\downarrow}}$}} & 44.91\smash{\rlap{$^{\textcolor{red}{26\%\downarrow}}$}} & 34.34\smash{\rlap{$^{\textcolor{red}{29\%\downarrow}}$}} \\
    T3D~\cite{zhang2026t3d}   & 48.65\smash{\rlap{$^{\textcolor{red}{7\%\downarrow}}$}} & 51.84\smash{\rlap{$^{\textcolor{red}{0\%\downarrow}}$}} & 4.15\smash{\rlap{$^{\textcolor{red}{8\%\downarrow}}$}} & 47.26\smash{\rlap{$^{\textcolor{red}{7\%\downarrow}}$}} & 39.06\smash{\rlap{$^{\textcolor{red}{10\%\downarrow}}$}} & 56.13\smash{\rlap{$^{\textcolor{red}{0\%\downarrow}}$}} & 50.59\smash{\rlap{$^{\textcolor{red}{2\%\downarrow}}$}} & 3.56\smash{\rlap{$^{\textcolor{red}{7\%\downarrow}}$}} & 57.51\smash{\rlap{$^{\textcolor{red}{5\%\downarrow}}$}} & 45.96\smash{\rlap{$^{\textcolor{red}{5\%\downarrow}}$}} \\
    \oursblkeight   & 50.89\smash{\rlap{$^{\textcolor{darkgreen}{3\%\downarrow}}$}} & 48.63\smash{\rlap{$^{\textcolor{darkgreen}{7\%\downarrow}}$}} & 4.37\smash{\rlap{$^{\textcolor{darkgreen}{3\%\downarrow}}$}} & 48.45\smash{\rlap{$^{\textcolor{darkgreen}{5\%\downarrow}}$}} & 42.77\smash{\rlap{$^{\textcolor{darkgreen}{2\%\downarrow}}$}} & 55.25\smash{\rlap{$^{\textcolor{darkgreen}{2\%\downarrow}}$}} & 47.34\smash{\rlap{$^{\textcolor{darkgreen}{8\%\downarrow}}$}} & 3.63\smash{\rlap{$^{\textcolor{darkgreen}{5\%\downarrow}}$}} & 57.50\smash{\rlap{$^{\textcolor{darkgreen}{5\%\downarrow}}$}} & 46.88\smash{\rlap{$^{\textcolor{darkgreen}{3\%\downarrow}}$}} \\
    \oursblkfour   & \textbf{52.79}\smash{\rlap{$^{\textcolor{darkgreen}{0\%\downarrow}}$}} & \textbf{50.27}\smash{\rlap{$^{\textcolor{darkgreen}{3\%\downarrow}}$}} & \textbf{4.43}\smash{\rlap{$^{\textcolor{darkgreen}{2\%\downarrow}}$}} & \textbf{49.62}\smash{\rlap{$^{\textcolor{darkgreen}{3\%\downarrow}}$}} & \textbf{43.54}\smash{\rlap{$^{\textcolor{darkgreen}{0\%\downarrow}}$}} & \textbf{55.46}\smash{\rlap{$^{\textcolor{darkgreen}{2\%\downarrow}}$}} & \textbf{48.74}\smash{\rlap{$^{\textcolor{darkgreen}{5\%\downarrow}}$}} & \textbf{3.66}\smash{\rlap{$^{\textcolor{darkgreen}{4\%\downarrow}}$}} & \textbf{58.29}\smash{\rlap{$^{\textcolor{darkgreen}{4\%\downarrow}}$}} & \textbf{47.61}\smash{\rlap{$^{\textcolor{darkgreen}{2\%\downarrow}}$}} \\
    \bottomrule
\end{tabular}%
  }
\end{table*}

\section{Experimetal Details}
\label{sec:appendix-experiment}
\subsection{Data}
\label{sec:appendix-data-details}
We aggregate reports from MIMIC-CXR~\cite{johnson2019mimic}, CheXpert-Plus~\cite{irvin2019chexpert}, ReXGradient~\cite{zhang2025rexgradient}, and IU-Xray~\cite{demner2016preparing} to form a unified corpus for CXR report generation, and apply a five-stage preprocessing pipeline to ensure consistency and training stability across sources.
\begin{enumerate}
    \item \textbf{Modality filtering.}
    we remove non-CXR studies and samples with incomplete clinical descriptions using clinical entity extraction.

    \item \textbf{Report standardization and normalization.}
    We use a prompt-based rewriting pipeline with BaichuanM2-32B to standardize clinical terminology, enforce a structured Findings--Impression format with JSON output, and normalize each report by explicitly enumerating all negative findings in order.

    \item \textbf{Semantic deduplication.}
    we embed reports with Qwen3-Embedding-8B~\cite{yang2025qwen3} and remove near-duplicates using a cosine-similarity threshold.
    After deduplication, the corpus contains approximately 260k reports from MIMIC-CXR, 250k from CheXpert-Plus, 190k from ReXGradient, and 30k from IU-Xray.

    \item \textbf{Bilingual augmentation.}
    we randomly sample 50\% of the original reports and translate them from English into Chinese using BaichuanM2-32B.
    Specific prompts are used to preserve medical terminology and report structure.

    \item \textbf{Supplemental general multimodal data.}
    we incorporate LLaVA-ReCap-558k~\cite{lillava} to preserve general vision-language grounding throughout the training process.
\end{enumerate}
This pipeline yields a large-scale, standardized, and bilingual CXR report corpus.

\subsection{Training details}
\label{sec:appendix-training-details}
All training stages are conducted on Alibaba PPUs.
To balance the information retention of high-resolution chest X-ray images and GPU memory consumption, we enforce a maximum pixel constraint of 2,250,000 throughout all stages.

\noindent\textbf{Stage 1 (AR continual pretraining).}
We perform full-parameter SFT on the base model Lingshu~\cite{xu2025lingshu}.
We employ the AdamW optimizer with a learning rate of $1\times 10^{-5}$, weight decay of 0.01, and a linear warmup over the first 0.03 epochs.

\noindent\textbf{Stage 2 (Diffusion adaptation).}
To accommodate the GPU memory overhead introduced by the extended vision token sequences of high-resolution chest X-ray images, we adopt a neat-packing training strategy.
This stage uses approximately 2.2\% of the Stage~1 SFT data, with the vision encoder and projector frozen and only the LLM backbone trained.

\noindent\textbf{Stage 3 (Distillation).}
We randomly sample 2.3\% of the SFT dataset as the distillation corpus.
This stage also employs neat-packing-based training, with the vision encoder and projector kept frozen.

\subsection{Metric Details}
\label{sec:metric_details}

We evaluate generated chest X-ray reports across three dimensions using five established metrics.
The calculation procedure and any task-specific adaptations for each metric are described below.

\noindent\textbf{ROUGE-L}~\cite{lin2004rouge} measures surface-level text overlap between the generated report and the reference.
It computes the longest common subsequence (LCS), capturing sentence-level structure and word-order similarity.

\noindent\textbf{CIDEr}~\cite{vedantam2015cider} evaluates report quality by emphasizing rare, clinically informative terms.
It computes the cosine similarity between TF-IDF-weighted $n$-grams of the generated and reference reports, giving greater weight to infrequent but diagnostically relevant terms over common stop words.

\noindent\textbf{Modified RaTEScore}~\cite{zhao2024ratescore} measures clinical entity-level accuracy.
A Named Entity Recognition (NER) module extracts medical entities and classifies them into five types: Anatomy, Abnormality, Disease, Non-Abnormality, and Non-Disease.
These entities are mapped to dense embeddings via a medical text encoder to resolve synonymy, and the final score is computed as a weighted F1 using a predefined clinical importance matrix.

However, directly applying the standard metric is not appropriate in our setting.
Since our normalized test set explicitly enumerates negative findings (see Sec.~\ref{sec:appendix-data-details}), conventional CXR report generation models would be unfairly penalized for not doing the same.
In clinical practice, what truly matters is whether a model correctly identifies positive pathologies.
We therefore set the importance weights of negative-related entity types (Non-Abnormality and Non-Disease) to zero, so that all models are evaluated solely on their ability to identify true positive findings.

\noindent\textbf{SemScore}~\cite{smit2020combining} evaluates semantic similarity between the generated and reference reports beyond exact lexical overlap.
It encodes both reports into dense embedding vectors via a pre-trained sentence transformer, and uses their cosine similarity as the evaluation score.

\noindent\textbf{Perplexity} quantifies the fluency of the generated text.
It is computed as the exponentiated average negative log-likelihood assigned to the token sequence by a standard causal language model.

In addition to quality metrics, Tab.~1 also reports two efficiency metrics to evaluate decoding speed.

\noindent\textbf{TPF} (Tokens Per Forward pass) measures the theoretical speedup relative to the AR baseline.
It is computed as the number of decoded tokens divided by the number of model forward passes during decoding.

\noindent\textbf{TPS} (Tokens Per Second) measures the practical decoding throughput.
It is computed as the number of tokens generated during the decoding stage divided by the elapsed time.

\subsection{Prompt templates}
\label{sec:appendix-prompts}
\begin{figure*}[t]
    \centering
    \includegraphics[width=0.8\linewidth]{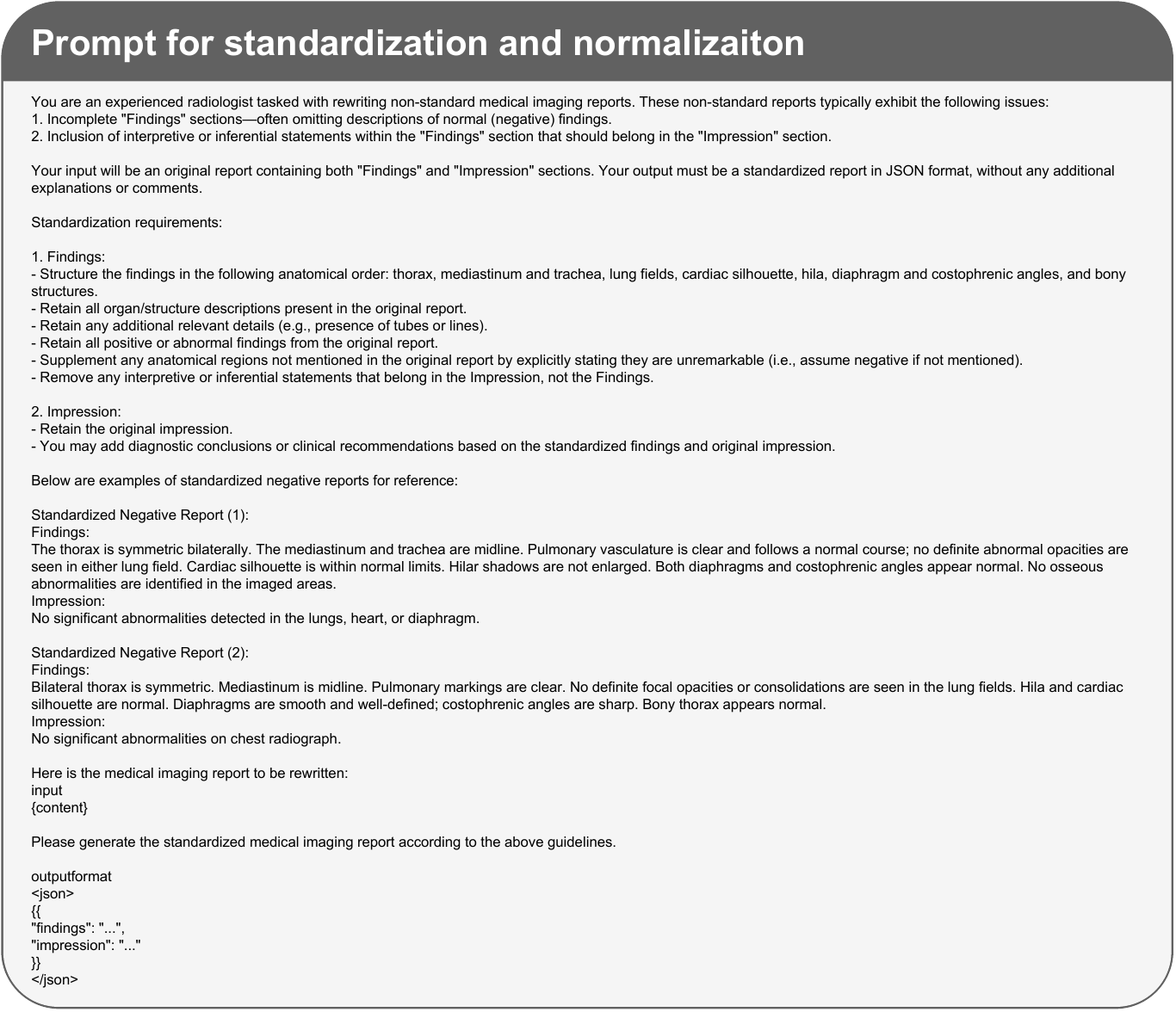}
    \caption{Prompt template for report standardization and normalization.
    The prompt instructs the model to rewrite a raw radiology report by: (1) separating the content into \textit{Findings} and \textit{Impression} sections with standardized clinical terminology, (2) explicitly enumerating all negative findings for clinical completeness, and (3) returning the result in a structured JSON format.}
    \label{fig:prompt-template-rewrite}
\end{figure*}

\begin{figure*}[t]
    \centering
    \includegraphics[width=0.8\linewidth]{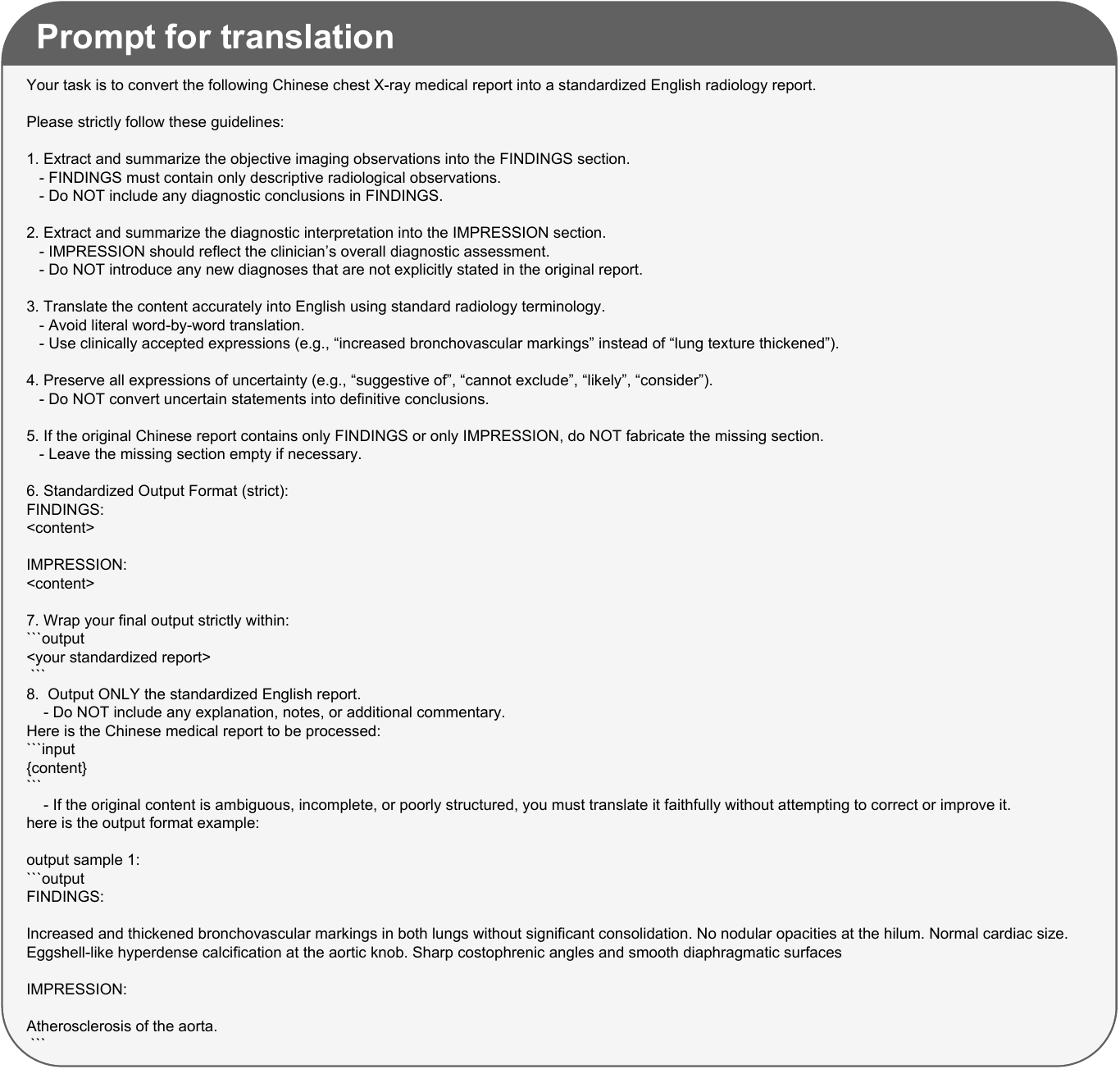}
    \caption{Prompt template for translating Chinese radiology reports into English, used to prepare evaluation data for the RaTEScore and SemScore metrics.}
    \label{fig:prompt-template-translation}
\end{figure*}

We provide the prompt templates used at each stage of our study below.
Fig.~\ref{fig:prompt-template-rewrite} shows the template for report structural standardization and report normalization.
Fig.~\ref{fig:prompt-template-translation} shows the templates for translating Chinese reports into English, as required by the RaTEScore and SemScore evaluation pipelines.
For both inference and evaluation, we use the following unified instruction prompt: \texttt{``Review this chest X-ray and write a report. Use this format: Findings: \{\}, Impression: \{\}.''}

\section{Implementation of baselines}
\label{sec:appendix-baselines}
\noindent\textbf{Autoregressive (AR) baselines.}
We adopt greedy decoding by setting the temperature to 0.
The maximum number of generated tokens is set to 512.
KV cache and FlashAttention are used to accelerate the decoding process.

\noindent\textbf{LLaDA-MedV.}
Following the default configuration of LLaDA-MedV~\cite{dong2025llada}, we employ a low-confidence remasking strategy with the response length set to $L=256$, the block length set to $B=64$, and the total number of sampling steps set to $Z=256$.
The decoding process is further accelerated using its default \texttt{fast\_dllm} configuration for efficient inference.

\noindent\textbf{Other distillation baselines.}
To further compare the efficiency of DCD, we implement several additional distillation baselines, all built upon the same base model \mbox{\textbf{\texttt{ECHO-Base}}\textsubscript{blk8}} and training data.
To enable direct comparison of inference speedup, all baselines adopt the semi-autoregressive decoding scheme aligned with BD3LM~\cite{arriolablock}.
We categorize them into three groups:
\begin{itemize}
  \item \textbf{Trajectory-based distillation.}
  Methods like dParallel~\cite{chen2025dparallel} directly fit the predictions of the current noisy block to pseudo-labels produced by multi-step denoising via cross-entropy loss.
  Following its official implementation, we apply both a CE loss and an Entropy Minimization loss, which encourage trajectory self-consistency and output confidence, respectively.
  We set \texttt{dparallel\_entropy\_weight} to 2.0 and \texttt{dparallel\_temperature} to 0.5.
  While this approach maintains trajectory consistency, directly applying CE loss does not resolve the mean-field bias problem.

  \item \textbf{Distribution-based distillation.}
  Methods like CD4LM~\cite{liang2026cd4lm} align the predicted distribution of the same block at a higher noise level (more tokens masked) with its distribution at a lower noise level (fewer tokens masked), using KL divergence as the alignment objective.
  Following its official implementation, we implement Discrete-Space Consistency Distillation using a combination of CE loss and KL divergence.
  We set \texttt{cdlm\_lambda} to 0.7 and \texttt{cdlm\_temperature} to 2.0.

  \item \textbf{Optimization-based distillation.}
  Methods like T3D~\cite{zhang2026t3d} and d3LLM~\cite{qian2026d3llm} optimize the model's predicted distribution on the multi-step teacher pseudo-label to correct the mean-field bias.
  Specifically, T3D adopts a DPO-style training objective that rewards distributions collected on multi-step teacher pseudo-labels, while penalizing distributions produced by the model's own few-step predicted labels.
  For both methods, we implement their training loss combinations following their original codebases.
\end{itemize}

\section{Detailed results}
\label{sec:appendix-detailed-results}
\noindent\textbf{Per-dataset results.}
Tab.~\ref{tab:performance_language_split_chexpertplus}, Tab.~\ref{tab:performance_language_split_rexgradient}, and Tab.~\ref{tab:performance_language_split_mimic} present a detailed per-metric comparison of our method against proprietary general models, autoregressive medical models, and other diffusion-based methods on each test set.
\base achieves state-of-the-art performance across all datasets, and \ours further maintains this performance with minimal quality degradation while reaching the maximum inference speedup.

\section{Fused Block KV Cache Analysis}
\label{sec:proof_fused_kv}
We show that Fused Block KV Cache preserves the total FLOPs of Vanilla Block KV Cache while halving the number of forward passes from $2N$ to $N$ for one-step-per-block decoding.

\noindent\textbf{Notation.}
Let $P$ denote the prompt length (including vision and instruction tokens), $B$ the block size, and $N$ the number of response blocks.
For a Transformer forward pass processing $q$ query tokens where each token attends to $\ell$ context positions, the FLOPs can be written as $F(q, \ell) = q \cdot g(\ell)$, where $g(\ell)$ is the per-token cost at context length $\ell$.
This linear decomposition in $q$ holds because all major Transformer operations (QKV projection, attention, output projection, and FFN) scale linearly with the number of query tokens.

\noindent\textbf{Vanilla Block KV Cache.}
For each block $n$ ($n = 0, \dots, N{-}1$) to be decoded, the KV cache holds $P + nB$ entries from the prompt and all previously decoded blocks.
Two separate forward passes are performed:
(i)~\textit{Denoise}: $q{=}B$ masked tokens, each attending to $P + nB + B = P + (n{+}1)B$ positions.
(ii)~\textit{KV Update}: $q{=}B$ decoded tokens, each attending to $P + nB + B = P + (n{+}1)B$ positions, to compute and cache their key-value states.
This yields $2N$ forward passes in total.

\noindent\textbf{Fused Block KV Cache.}
Block $0$ is handled with a standard denoise forward ($q{=}B$), and its KV update is deferred.
For each subsequent block $n$ ($n = 1, \dots, N{-}1$), the deferred KV update from block $n{-}1$ and the denoise of block $n$ are merged into one fused forward ($q{=}2B$).
This yields $N$ forward passes in total.

\noindent\textbf{FLOPs equivalence.}
Each fused forward for block $n$ ($n \geq 1$) replaces exactly two Vanilla operations: the KV update of block $n{-}1$ and the denoise of block $n$.
At this point, the KV cache holds $P + (n{-}1)B$ entries, since the KV update for block $n{-}1$ has not yet occurred.
Under the block-causal attention mask, each token in the fused forward attends to the same context as in the corresponding Vanilla pass:
\begin{itemize}[nosep]
\item The $B$ decoded tokens from block $n{-}1$ attend to $P + (n{-}1)B + B = P + nB$ positions, matching the Vanilla KV update for block $n{-}1$.
\item The $B$ masked tokens for block $n$ attend to $P + (n{-}1)B + B + B = P + (n{+}1)B$ positions, matching the Vanilla denoise for block $n$.
\end{itemize}
The combined FLOPs of the two Vanilla passes are:
\begin{equation}
F_{\text{vanilla}} = B \cdot g(P + nB) + B \cdot g\bigl(P + (n{+}1)B\bigr)\,.
\end{equation}
Since the fused forward processes the same $2B$ tokens with identical per-token contexts, its FLOPs are:
\begin{equation}
F_{\text{fused}} = B \cdot g(P + nB) + B \cdot g\bigl(P + (n{+}1)B\bigr) = F_{\text{vanilla}}\,.
\end{equation}
This equality holds for every $n \geq 1$.
Therefore, Fused Block KV Cache introduces no additional FLOPs while reducing the number of forward passes from $2N$ to $N$, directly lowering inference latency.

\end{document}